%% file: main.tex
\documentclass{article}

\usepackage{arxiv}

\usepackage[utf8]{inputenc} 
\usepackage[T1]{fontenc}    
\usepackage{hyperref}      
\usepackage{url}           
\usepackage{booktabs}      
\usepackage{amsfonts}      
\usepackage{nicefrac}    
\usepackage{microtype}    
\usepackage{lipsum}
\usepackage{graphicx}
\usepackage{wrapfig}
\usepackage[backend=biber, 
    style=apa]{biblatex}
\usepackage{amsmath}
\usepackage[short]{optidef}

\usepackage[toc,page]{appendix}
\usepackage{amsthm}
\usepackage{subcaption}
\DeclareMathOperator{\Tr}{tr}
\DeclareMathOperator{\sgn}{sgn}
\DeclareMathOperator{\symlog}{symlog}
\usepackage{tikz}
\usetikzlibrary{arrows, automata}
\usetikzlibrary{bayesnet}
\usepackage{dirtytalk}

\title{Slow Feature Analysis on Markov Chains\\from Goal-Directed Behavior}

\author{
 Merlin Schüler\thanks{
  Institute for Neural Computation, Faculty of Computer Science, Ruhr University Bochum,  Germany.\\Corresponding author:\texttt{merlin.schueler@ini.rub.de}}
   \And
 Eddie Seabrook\footnotemark[1]
  \And
 Laurenz Wiskott\footnotemark[1]
}

\begin{document}
\maketitle
\begin{abstract}
Slow Feature Analysis is a unsupervised representation learning method that extracts slowly varying features from temporal data and can be used as a basis for subsequent reinforcement learning.
Often, the behavior that generates the data on which the representation is learned is assumed to be a uniform random walk. Less research has focused on using samples generated by goal-directed behavior, as commonly the case in a reinforcement learning setting, to learn a representation. In a spatial setting, goal-directed behavior typically leads to significant differences in state occupancy between states that are close to a reward location and far from a reward location. 

Through the perspective of optimal slow features on ergodic Markov chains, this work investigates the effects of these differences on value-function approximation in an idealized setting. Furthermore, three correction routes, which can potentially alleviate detrimental scaling effects, are evaluated and discussed. In addition, the special case of goal-averse behavior is considered.
\end{abstract}

\section{Introduction}
The learning of representations is a key challenge in machine learning, as it can facilitate faster learning of downstream tasks by increasing data efficiency without manual feature engineering. Furthermore, good representations are often task-agnostic and domain-specific and can thus be transferred to multiple tasks in related domains, thus allowing subsequent learning to focus on the task itself.

In Slow Feature Analysis (SFA) \parencite{Wiskott1998, WiskottSejnowski2002}, a series of mappings $g_i$ from the samples to the low-dimensional representation learned so that they optimize
\begin{mini!}   
  {g_i}{\big<(g_i\left(\mathbf{x}_{t+1}) - g_i(\mathbf{x}_{t})\right)^2\big>_t}{}{}
  \addConstraint{\big<g_i(\mathbf{x}_t)\big>_t=0}{}
  \addConstraint{\big<g_i(\mathbf{x}_t)g_j(\mathbf{x}_t)\big>_t=0,}{}{\quad\forall j < i}  
  \addConstraint{\big<g_i(\mathbf{x}_t)^2\big>_t=1,}{}{\quad\forall i}
\label{eq:sfa_optimization_problem}
\end{mini!}
\noindent
where $\big< \cdot \big>_t$ is the average over time. Solving this optimization problem leads to a set of mappings, ordered by their respective slowness.

Although some frameworks, such as representation policy iteration \parencite{Mahadevan2012}, account for the updating of the representation during later stages of behavior learning, a representation is commonly learned and fixed before any task-specific learning occurs. In case of SFA, this means collecting samples from a random walk until the representation is stable, while discarding any task-specific reward that an environment might provide. The specific random walk used to generate SFA features is the object of investigation in this work.

A recent approach by \textcite{Hakenes2019} to combine task-specific learning, in this case reinforcement learning, with end-to-end slowness optimization through gradient-based SFA \parencite{Schueler2019} have yielded negative results: Despite its efficacy as a pre-learned representation, the results of using SFA for augmentation were insignificant at best and detrimental at worst. A deeper analysis of these effects has not yet been published, which we attribute partially to a gap in understanding between slow features that are derived from random walks and slow features derived from directed behavior which occurs often in late stages of reinforcement learning. This work partially addresses this issue by investigating slow representations learned from goal-directed behavior in spatially connected environments.

Section \ref{sec:mcsfa:background} establishes some formal background, followed by the proposal of a natural analogue of SFA on stochastic processes in Section \ref{sec:mcsfa}. 
The analysis is focused on Markov chains for multiple reasons: They provide a simple model for directed behavior, slowness is typically defined over one-step transitions, and the established results connect well with Markov Decision Processes, which are the theoretical underpinning of an overwhelming part of reinforcement learning research. 

The derived formulation and its optimal solutions integrate well with known spectral embeddings for directed graphs \parencite{chung2005, johns2007}. In Section \ref{subsec:weakly}, the optimal solutions are visually inspected for simple Markov chains with respect to qualitative differences, and correction mechanisms are proposed. This informs a quantitative analysis in Section \ref{sec:mcsfa_approximation}, which focuses on regression performance in value-function approximation in spatially connected environments.

We conclude with a discussion of the results, concrete questions for further research, as well as suggestions for possible improvements in Section \ref{sec:discussion}.

\section{Related Work}
\label{sec:mcsfa_related_work}
Despite the discrepancy between the investigation of undirected versus directed behavior for slowness extraction, there is a rich body of research that gives this work context. 

\paragraph{Laplacian eigenmaps and SFA} \textcite{Sprekeler2009}  used probabilistic formalism by assuming an ergodic time-series as input to SFA. This implies a probability density on a manifold in input space as well as its time-derivative. In later work, the author used that formalism to establish a connection between a generalized version of SFA and Laplacian eigenmaps \parencite{Sprekeler2011}.
\paragraph{SFA and Markov chains} 
\textcite{Klampfl2013} proposed the construction of a Markov chain from labeled training data and demonstrated that slow features learned from a time-series generated by this chain can be used for supervised classification. Later, \textcite{Escalante2013} built on this by proposing graph-based SFA, a method for representation learning on training data in which data points are arranged in a graph, and showed that it is equivalent to applying SFA to the Markov chain induced by a random walk on this training graph and that it can also be used as effective representation for subsequent supervised classification. Graph-based SFA is strongly related to generalized SFA as proposed in \textcite{Sprekeler2011} as well as the construction presented in this thesis. 

\paragraph{Environments with directed transitions} Proto-value functions for spatial environments with directed transitions have been investigated by \textcite{johns2007} using a graph symmetrization proposed by \textcite{chung2005}, which notably coincides with the derivation of SFA on ergodic Markov chains. They conclude that the symmetrization can account for one-way transitions in environments\footnote{By enforcing ergodicity.}, but do not investigate the effect of goal-directed behavior and the stationary distribution or state occupancy on the extracted features.

\paragraph{Optimal slow features in spatial environments} \textcite{Franzius2007} derive theoretically optimal features for random behavior in spatial environments. In this derivation, they identify a dependency of the feature amplitude on state occupancy, which is the same effect discussed in this work. However, the consequences of this, for example, when using SFA as basis functions, are not discussed in detail.

\paragraph{Probabilistic SFA} \textcite{Turner2007} determined that the solutions to linear SFA coincide with the maximum-likelihood solution of a latent Gaussian dynamical system when observed after a linear transformation. PSFA is related to the research presented in this work mainly through the use of probabilistic formalism and by assuming the Markov property on the latent variables, but assumes a more specific family of Markov chain and a linear relationship between input and representation. 

\paragraph{Directed SFA} In a thorough and rigorous theoretical treatment, \textcite{Boehmer2013} identified equivalencies for general Markov chains and the symmetrization of the transition dynamics, including directed versions thereof, induced by SFA. The formulation used is in line with the research mentioned above as well as with the one used in this work. In addition to the work mentioned above, they highlight a specific dependence of SFA features on the stationary distribution as does the work presented in this thesis. However, their discussion is specifically aimed at the relative difference in features induced by a mixture of latent generative factors, i.e., change in orientation and change in position.\footnote{They also propose a correction mechanism, which corresponds to \textit{learning rate adaptation} as proposed by \textcite{Franzius2007}}. Specifically, the characteristics of slow features relating to local occupancy, as are the focus of this work, were not discussed.


\section{Reinforcement Learning and Markov Decision Processes}
\label{sec:mcsfa:background}
In machine learning, the field that deals with environments and behavior is reinforcement learning (RL). Within RL, the formal language used to describe environments are \textit{Markov decision processes} of the form
\begin{equation}
    \mathcal{M} = \{\mathcal{S}, \mathcal{A}, \mathcal{R}, \mathcal{T}\}
\end{equation}
with a finite state space $\mathcal{S}$, a finite action space $\mathcal{A}$, a reward function $\mathcal{R}(s)$ that assigns each state $s$ a reward that an agent receives upon transitioning from\footnote{Multiple definitions are possible, depending on the context, that take the destination state, goal state, action or any subset into account.} $s$, and transition dynamics $\mathcal{T}(s'|s,a)$ that determines the probability of transitioning into a state $s'$ given a state $s$ and a chosen action $a$.

The behavior of an agent acting in a Markov decision process is defined by a probability distribution $\pi(a|s)$, called a \textit{policy}, which expresses the probability of selecting an action $a$ when in state $s$. The policy $\pi$ and together with the transition dynamics $\mathcal{T}$ induce a Markov chain with transition probabilities $\mathcal{P}(s'|s) = \sum_a \mathcal{T}(s'|s,a) \pi(a|s)$. For explicit states $s_u$ and $s_v$, we denote $P_{uv}$ as the probability $\mathcal{P}(s'=s_v|s=s_u)$. If the Markov chain has a stationary distribution, this is denoted as $\mu$. Different policies induce different Markov chains and, when relevant, the policy used is indicated by superscript as $\mathcal{P}^\pi$, $P^\pi$ and $\mu^\pi$.

\paragraph{Value functions} The canonical objective in reinforcement learning is the maximization of the (expected) collected reward over time. This objective is often expressed and evaluated through value-functions, which allow to compare different behaviors when assuming to start from a certain state $s$ (possibly a first action $a$) and from there following a policy.

Formally, when executing an action $a$ in state $s$ and subsequently following the given policy $\pi$, the state-action value-function is defined as
\begin{equation}
    Q^\pi(s, a) = \mathbb{E}_{s_t \sim P^\pi}\left[\sum_{t=1}^\infty \gamma^{t - 1} \mathcal{R}(s_{t})|s_1=s, a_1=a\right]
\end{equation}
or, if $a$ is also distributed according to $\pi$, one can consider the state value-function
\begin{equation}
    V^\pi(s) = \mathbb{E}_{s_t \sim P^\pi}\left[\sum_{t=1}^\infty \gamma^{t - 1} \mathcal{R}(s_{t})|s_1=s\right].
\end{equation}
The discount factor $\gamma \in [0, 1)$ expresses preference for myopic or farsighted behavior, and, aligning with similar studies, $\gamma = 0.95$ is considered in the following sections. If either value-function corresponds to an optimal policy, it is denoted as $V^*$ or $Q^*$.

\paragraph{Approximation} In practice, value-functions are estimated from data and approximated by a regression model. The specific details of the estimation and approximation are somewhat independent design decisions in many modern reinforcement learning algorithms. For example, in Deep Q-Learning \parencite{mnih2013playing}, the approximation is performed by a deep neural network trained to minimize the mean square error between the network output and $Q^*$. However, since $Q^*$ cannot be evaluated directly, an estimate is produced using a standard reinforcement learning approach called \textit{Q-learning} \parencite{Watkins1992} which serves as an approximation target instead.

The efficacy of such an approach depends in large part on the regression model to quickly, robustly, and accurately learn the approximation target from the data. Consequently, the efficacy of a representation in the context of reinforcement learning can be judged simply by how well it supports such an approximation. The work investigates how well optimal SFA features serve as a basis for the approximation of $V^*$, but all results are assumed to transfer to the approximation of $Q^*$. The following assumptions and idealizations are made:
\begin{itemize}
    \item The approximation target $V^*$ is not estimated, but provided as ground truth. This excludes unrelated sources of approximation error, such as bootstrapping bias in Q-learning.
    \item The regression model is linear and of the form \begin{align}
        \hat{V}(s) = \mathbf{w}^T \mathbf{g}(s)
    \end{align} with parameters $\mathbf{w}$ and a vectorial slow feature representation $\mathbf{g}(s)$ of a state $s$. The parameters are identified via ordinary least squares.
    \item The learning of the approximation is not based on samples, but instead leverages the stationary distribution of the environment. This can also be seen as an infinite-sample case.
    \item The distribution over states is induced by an exploratory behavior policy, while the approximation target $V^*$ corresponds to an optimal policy. This is the most common setting in reinforcement learning, known as \textit{off-policy learning} \parencite{Sutton98}.
\end{itemize}

\paragraph{Behavior} Reinforcement learning depends on behavior in the form of a policy $\pi$ to generate samples. A purely exploratory policy is often inefficient, while a purely exploitative policy is only sensible when sufficient knowledge about the environment is incorporated. Thus, policies are often defined as a trade-off between exploitation and exploration.

The most common form of policy is the $\varepsilon$\textit{-greedy policy} \parencite{Sutton98}. In a state $s_t$, the agent picks the optimal action\footnote{Or the current best estimate thereof.\label{footnote:qstar}} ($a^* ={\arg \max}_a Q^*(s_t,a)$) with probability $1-\varepsilon$ and a random action with probability $\varepsilon$. A variant used in this work picks the optimal action with probability $1-\zeta$ and a nonoptimal action with probability $\zeta$ and is hence called the $\zeta$ -greedy policy. The difference is subtle, but leads to qualitatively different behaviors: While $\varepsilon=1$ leads to a uniform policy, $\zeta=1$ leads to a distinctly suboptimal policy. This allows for the investigation of goal-averse behavior in the following sections.

Another way to include exploration is the use of a \textit{Boltzmann policy} \parencite{Sutton98, Szepesvari2010}. It is widely used, although less common than $\varepsilon$-greedy. Here, the probability of selection for each action $a$ in a state $s$ depends on the value $Q^*(s, a)$\footnote{Or the current best estimate thereof.}. The probability distribution of the policy is constructed using the softmax over all possible actions as 
\begin{equation}
    \pi(a|s) = \frac{e^{\beta Q^*(s,a)}}{\sum_i e^{\beta Q(s, a_i)}}
\end{equation}
where $\beta$ controls the goal-directedness of the exploration. Typically, $\beta > 0$, but we also consider the cases $\beta=0$ (uniform behavior) and $\beta < 0$ (goal-averse behavior). When the difference in value between optimal and nonoptimal actions is large, the action selection is decisive. When the difference is small, the selection probability is distributed more evenly among actions.


\section{SFA on Markov Chains}
\label{sec:mcsfa}
In this section, SFA is formulated on stochastic processes in general and the special case of Markov chains is derived in particular. As outlined in the previous section, such a Markov chain could be the result of Markov decision process and a policy.

Assuming a discrete-time stochastic process $\mathbb{S}=\{s_t\}_{t\in\mathbb{N}}$ where each individual $s$ lives in a finite state space $s_t\in\mathcal{S}$ and each sample $S \sim \mathbb{S}$ corresponds to an instantiation of the process and thus to an infinite-length time-series. For a (bounded) function $g$ that acts on the state space, we denote $g(S)$ as an element-wise application of this function to all members of a sample. The slowness of given $g$ on a sample $S$ is consequently defined as
    \begin{equation}
    \Delta(g(S)) = \lim_{T\rightarrow \infty} \frac{1}{T-1}\sum_{t=1}^{T} (g(s_{t+1}) - g(s_t))^2.
    \end{equation}
    
From this definition on samples, the objective on a process $\mathbb{S}$ can be defined
as the expected slowness of its samples. Similarly, analogues of the SFA constraints can be defined for a process, leading to the following optimization problem on $g_i$:
\begin{figure}[h]
\vspace*{-4mm}
\begin{mini!}
  {g_i}{\mathbb{E}_{S\sim\mathbb{S}}\big[\Delta(g_i(S))\big]}{\label{eq:mcsfa_optimization_problem}}{} \label{eq:mcsfa_objective}
  \addConstraint{\mathbb{E}_{S\sim\mathbb{S}}\Big[\lim_{T\rightarrow\infty} \frac{1}{T-1} \sum_{t=1}^{T} g_i(s_t)\Big]}{=0}{} \label{eq:mcsfa_constraint_zeromean}
  \addConstraint{\mathbb{E}_{S\sim\mathbb{S}}\Big[\lim_{T\rightarrow\infty} \frac{1}{T-1} \sum_{t=1}^{T} g_i(s_t)g_j(s_t)\Big]}{ = \delta_{ij},\quad}{\forall j \leq i} \label{eq:mcsfa_constraint_unitvariance}
\end{mini!}
\vspace*{-4mm}
\end{figure}

\noindent where the roles of the constraints are (in expectation) equivalent to those of the original SFA formulation, i.e., to avoid constant or redundant solutions.

Up to this point this is as a general definition and neither implies the actual existence of each limit and nor is it universally. The results of this work are restricted to a particular family of processes:
\textbf{ergodic Markov chains on a finite state-spaces}. This limitation might appear drastic at first, however, Markov processes on finite state-spaces are prevalent even in modern reinforcement learning. Ergodicity is a stronger assumption, but arises naturally for spatially-connected environments with some (arbitrarily small) amount of random exploration, except for specific cases (e.g.\ , rooms that can be entered but not exited).

For Markov chains of this type, it is well-known that they possess a limiting distribution
\begin{align}
    \forall s_u, s_v:\quad &\lim_{t\rightarrow\infty} \mathcal{P}(s_{t}=s_u|s_{1}=s_v)\\ = &\lim_{t\rightarrow\infty} \mathcal{P}(s_{t}=s_u)= \mu_u
\end{align}
independent of starting state $s_1$. This distribution is called the \textit{stationary distribution} of the process.


For any sample of the process, $\mu_u$ and  $\mu_u P_{uv}$  capture the fraction of visits to the state $s_u$ and fraction of transitions $s_u\rightarrow s_v$, respectively \parencite{bertsekas2002}. As none of the quantities in optimization problem \eqref{eq:mcsfa_optimization_problem} depends on the order of terms, this leads to a closed form for these limits. Since these state-visitation and transition frequencies are the same for every sample, the expectation can be dropped, resulting in a simplified optimization problem with the objective
\begin{align}
&\sum_{u, v} \mu_{u}P_{uv}(g_i(s_u) - g_i(s_v))^2 \label{eq:prob_sum_objective}\\
=& \sum_{u, v} M_{uv}(g_i(s_u) - g_i(s_v))^2 
\end{align}
with coefficients $M_{uv} = \frac{1}{2}(\mu_{u}P_{uv} + \mu_{v}P_{vu}) = M_{vu}$ due to the symmetry of the squared difference. From standard marginalization follows
\begin{equation}
    \sum_u M_{vu} = \sum_u M_{uv} = \mu_v
\end{equation}
allowing to reformulate the objective function for finding optimal function values $y_{iu} = g_i(s_u)$ and consequently the optimization problem as follows
\begin{figure}[h]
\vspace*{-6mm}
\begin{mini!}   
  {\mathbf{y}_i}{\sum_{u, v} M_{uv}(y_{iu} - y_{iv})^2 = 2 \mathbf{y}_i^T(\mathbf{D} - \mathbf{M})\mathbf{y}_i\propto \mathbf{y}_i^T(\mathbf{D} - \mathbf{M})\mathbf{y}_i}{}{} \label{eq:mcsfa_objective_simplified}
  \addConstraint{\sum_{u}\mu_{u} y_{iu}=\mathbf{y}_i^T\mathbf{D}\mathbf{1}}{=0}{}  \label{eq:mcsfa_constraint_zeromean_simplified}
  \addConstraint{\sum_{u}\mu_{u} y_{iu}y_{ju} = \mathbf{y}_i^T\mathbf{D}\mathbf{y}_j}{ = \delta_{ij},}{\quad\forall j \leq i} \label{eq:mcsfa_constraint_unitvariance_simplified}
\end{mini!}
\label{eq:mcsfa_optimization_problem_simplified}
\vspace*{-10mm}
\end{figure}

\noindent where $\mathbf{D}$ is a diagonal matrix with entries $D_{vv} = \sum_u M_{uv} = \mu_v$. 

This is comparable in its approach to \textcite{Wiskott2003}, where optimal responses of continuous-time SFA are determined under the assumption that the output features are independent of the input signals (they are \textit{free} responses). 
This work employs the slightly different phrasing that optimal features are of interest that disregard any restriction (or definition) of the actual functional forms of $g_i$. Most spectral embedding methods disregard the notion of a functional form altogether in their derivation, although it can be added as an extension \parencite{Bengio04} to allow for out-of-sample embeddings. In both cases, this renders the analysis unconstructive in the sense that it does not provide explicit direction on how to find a mapping that produces such optimal output features, but in return allows for a qualitative investigation that is unconfounded by any assumed exact nature of such mapping.

A more common form of optimization problem can be produced by dropping the zero-mean constraint \eqref{eq:mcsfa_constraint_zeromean_simplified}. In that case, $\mathbf{y}_0 = \mathbf{1}$ becomes a globally optimal, but trivial, solution and any feasible $\mathbf{y}_{>0}$ are necessarily zero-mean due to the unit-variance constraint \eqref{eq:mcsfa_constraint_unitvariance_simplified}. 
\begin{figure}[!h]
\vspace*{-5mm}
\begin{mini!}   
  {\mathbf{y}_i}{\mathbf{y}_i^T(\mathbf{D} - \mathbf{M})\mathbf{y}_i}{\label{eq:mcsfa_optimization_problem_short}}{} \label{eq:mcsfa_objective_short}
  \addConstraint{\mathbf{y}_i^T\mathbf{D}\mathbf{y}_j}{ = \delta_{ij},}{\quad\forall j\leq i} \label{eq:mcsfa_constraint_unitvariance_short}
\end{mini!}
\vspace*{-10mm}
\end{figure}
In the following notions, $\mathbf{y}_0$ is contained in the derivation for ease of notation, but disregarded for any further discussion of the embedding.

It should also be noted that $(\mathbf{D} - \mathbf{M})$ is equivalent to a definition for a symmetrized Laplacian matrix of directed graphs used by \textcite{chung2005} and \textcite{johns2007} for directed proto-value functions. 

As a final step, the notion of order is discarded from the optimization problem, although it will naturally be reintroduced due to the nature of the solution. Instead of formulating it sequentially for individual $\mathbf{y}_i$, the unordered problem can therefore be written in terms of a single matrix $\mathbf{Y}=(\mathbf{y}_0, \cdots, \mathbf{y}_e)$, where $e$ is the feature dimensionality (the number of slow features):

\begin{figure}[!h]
\vspace*{-5mm}
\begin{mini!}   
  {\mathbf{Y}}{\Tr\left(\mathbf{Y}^T(\mathbf{D} - \mathbf{M})\mathbf{Y}\right)}{\label{eq:mcsfa_optimization_problem_matrix_short}}{} \label{eq:mcsfa_objective_matrix_short}
  \addConstraint{\mathbf{Y}^T\mathbf{D}\mathbf{Y}}{ = \mathbf{I}_{e+1}}{} \label{eq:mcsfa_constraint_unitvariance_matrix_short}
\end{mini!}
\vspace*{-10mm}
\end{figure}

While this form of optimization problem and its solutions are well-known, resources outlining the process of getting to the said solutions appear to be somewhat scarce, particularly outside optimization literature. This is why the process is illustrated in the following, using the method of Lagrange multipliers.

The corresponding Lagrange function can be written as
\begin{equation}
    \mathcal{L}(\mathbf{Y}, \mathbf{\Lambda}) = \Tr\left(\mathbf{Y}^T(\mathbf{D} - \mathbf{M})\mathbf{Y}\right) - \Tr\left(\mathbf{\Lambda}(\mathbf{Y}^T\mathbf{D}\mathbf{Y} - \mathbf{I}_{e+1})\right)
\end{equation}
\noindent where $\mathbf{\Lambda}$, w.l.o.g.\ , can be formulated as diagonal matrix with the Lagrange multipliers $\lambda_i$ as its diagonal elements\footnote{See \textcite{ghojogh2023eigenvaluegeneralizedeigenvalueproblems} for a good explanation why this is the case.}. To find candidates for optima, stationarity of the Lagrangian is assumed: 
\begin{align}
    &\frac{\partial \mathcal{L}}{\partial \mathbf{Y}} = \mathbf{0} \\
    \Leftrightarrow\quad&
    \frac{\partial \Tr\left(\mathbf{Y}^T(\mathbf{D} - \mathbf{M})\mathbf{Y}\right)}{\partial \mathbf{Y}} - \frac{\Tr\left(\mathbf{\Lambda}(\mathbf{Y}^T\mathbf{D}\mathbf{Y} - \mathbf{I}_{e+1})\right)}{\partial \mathbf{Y}} = \mathbf{0} \label{eq:trace_derivative_start}\\
   \Leftrightarrow& 2(\mathbf{D}-\mathbf{M})\mathbf{Y}- 2 \mathbf{D}\mathbf{Y}\mathbf{\Lambda} = \mathbf{0}\\
    \Leftrightarrow\quad& (\mathbf{D}-\mathbf{M})\mathbf{Y} = \mathbf{D}\mathbf{Y}\mathbf{\Lambda} \label{eq:generalized_eigenvalue_problem_matrix_form}
\end{align}

The matrix derivatives are provided in detail in Appendix \ref{appendix:matrix_derivatives}. Since the resulting equation \eqref{eq:generalized_eigenvalue_problem_matrix_form} describes a generalized eigenvalue equation, the feasible solutions to the optimization problem \eqref{eq:mcsfa_optimization_problem_matrix_short} are the generalized eigenvectors $Y_{i\cdot}$ as columns of $\mathbf{Y}$ with Lagrange multipliers being the corresponding eigenvalues $\lambda_i$.

This has the consequence that, for all feasible solutions, the objective function evaluates to
\begin{align}
    \Tr\left(\mathbf{Y}^T\smash{\underbrace{(\mathbf{D} - \mathbf{M})\mathbf{Y}}_{=\mathbf{D}\mathbf{Y}\mathbf{\Lambda}}}\right) = \Tr(\underbrace{\mathbf{Y}^T\mathbf{D}\mathbf{Y}}_{=\mathbf{I}_{e+1}}\mathbf{\Lambda}) = \Tr(\mathbf{\Lambda}) = \sum_{i=0}^e \lambda_i \label{eq:mcsfa_objective_as_eigenvalues}
\end{align}
\noindent The dimension $e+1$ corresponds to the number of columns in $\mathbf{Y}$. It is straightforward to see that the set of smallest eigenvalues will minimize the objective \eqref{eq:mcsfa_objective_as_eigenvalues} and thereby the corresponding eigenvectors are the optimal features. Note that this also implicitly reestablishes the ordering inherent to the sequential formulation eq. \eqref{eq:mcsfa_optimization_problem_short}.
Thus, one can solve the Markov chain formulation of SFA by solving a generalized eigenvalue problem and taking the $e$ eigenvectors corresponding to the smallest eigenvalues (discarding the trivial solution). The simulations in the remainder of this work are based on these solutions.

This is equivalent to Laplacian eigenmaps on a weighted undirected graph defined by the weight matrix $\mathbf{M}$, a connection that has been previously investigated by \textcite{Sprekeler2011}. 

\section{Features of Weakly-Directed Behavior}
\label{subsec:weakly}
To illustrate the influence of directed behavior on extracted slow features, first a simple Markov chain will be discussed which can be considered a simplified variant of a finite \textit{birth-death process}\footnote{The name stems from the fact that it is a simple population model for which the transition probabilities correspond to a member of the population either dying or being born.}. It can be represented as a finite linear graph with states $\{s_0, \dots, s_{N-1}\}$, as depicted in Figure \ref{fig:gd_graph}, and is parameterized by a single parameter $\theta$, which corresponds to the probability that a transition $s_i \to s_{\min(N-1, i+1)}$ occurs. Inversely, $1 - \theta$ corresponds to the probability that a transition $s_i \to s_{\max(0, i-1)}$ occurs.
\input{gdprocess}

\noindent This can be understood as a simple model of goal-directed behavior in which a tendency to the left (toward $s_0$) is expressed through $\theta < 0.5$, a tendency to the right (toward $s_{N-1}$) is expressed through $\theta > 0.5$, and no tendency (uniformly random behavior) is expressed as $\theta = 0.5$. For any $\theta \in (0,1)$, this Markov chain is ergodic and its stationary distribution can be found analytically and follows a geometric shape $\mu_i \propto (\frac{\theta}{1-\theta})^i$ \parencite{bertsekas2002}. 

\begin{figure}[ht]
    \centering
    \includegraphics[width=0.99\columnwidth]{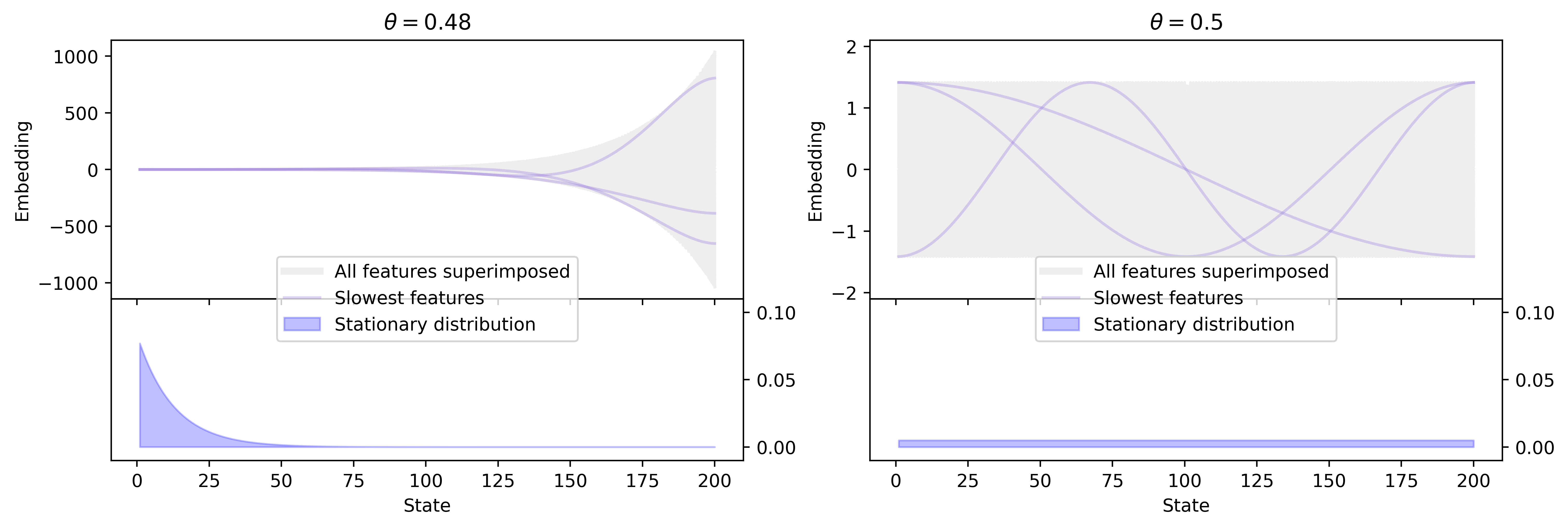}
    \vspace*{-4mm}
         \caption{The optimal embeddings for the birth-death-process with $N=200$ and corresponding stationary distributions for two settings of $\theta$. Slowest three features shown in purple, all other features superimposed in gray.}
    \label{fig:bdprocess_features}
\end{figure}

It is straightforward to formulate and solve the corresponding SFA optimization problem \eqref{eq:mcsfa_optimization_problem_matrix_short} to acquire the optimal slow features. In Figure \ref{fig:bdprocess_features}, these features are shown for two settings of $\theta$ with the three slowest features depicted in purple, whereas all other features are superimposed in gray. 

Setting $\theta=0.5$ results in a uniform process with a uniform stationary distribution. Such behavior would typically be used for the training of SFA and it results in "textbook" slow features, as often seen in the literature.

However, even a very slight deviation from such purely exploratory behavior, expressed by $\theta=0.48$, leads to a significant change: All resulting features are flat in areas of high occupancy, but scaled up in areas of lowest occupancy.
This is not an artifact of this particular process, but, in fact, a general dependency on the stationary distribution. This dependency is partially expressed through the constraint \eqref{eq:mcsfa_constraint_unitvariance_matrix_short}, which can also be written as
\begin{align}
    \mathbf{y}^T\mathbf{Dy} = \sum_i\mu_i y_i^2 = 1.
\end{align}
\noindent As all $\mu_i$ and $y_i^2$ are nonnegative, it holds that
\begin{align}
    \mu_i y_i^2 \leq 1\quad\text{and thus}\quad y_i \leq \pm \frac{1}{\sqrt{\mu_i}}
\end{align}
for each $y_i$ and $\mu_i$. This bound is tight only if all but one $y_i$ are $0$, but for any case in which part of the variance is distributed over a part of the states, the bound tightens for all others. Specifically, if some set of indices $\mathbb{J}_\text{fixed}$ has variance $v_\text{fixed}=\sum_{j\in \mathbb{J}_\text{fixed}} \mu_j y_j^2$, then the bound tightens to 
\begin{align}
    \mu_i y_i^2 \leq 1 - v_\text{fixed}\quad\text{and thus}\quad y_i \leq \pm \sqrt{1 - v_\text{fixed}}\frac{1}{\sqrt{\mu_i}} \label{eq:feature_bounds}
\end{align}
for all other $i\notin\mathbb{J}_\text{fixed}$.

This is not a formal proof that optimal features are generally impacted by such scaling and, in fact, the scaling is not only caused by the constraint but by the contribution of the stationary distribution in the objective function as well. However, since the slowness objective by definition promotes an even distribution of variance leading to a tightening of the bound for individual states, a general effect seems plausible and is, in fact, confirmed by all the experiments conducted.

\section{Correction Mechanisms}
\label{sec:mcsfa_correction_mechanisms}
The findings in Section \ref{sec:mcsfa} imply three correction routes, each of which can be applied at a different step in the feature extraction.  

The most straightforward intervention is a \textbf{behavior modification} at the time of sample collection. For example, if the behavior is generated through a $\zeta$-greedy or $\varepsilon$-greedy policy, increasing the amount of exploration will lead to a more even distribution of the stationary probability among states. This will lead to inefficiencies due to the oversampling of suboptimal actions. Furthermore, Section \ref{sec:mcsfa} shows that even slightly goal-directed behavior can exhibit a significant impact on the resulting slow features. A less drastic intervention is to prefer Boltzmann exploration, allowing for behavior similar to $\zeta$-greedy or $\varepsilon$-greedy policies when close to a reward, but a more even distribution of stationary probability overall by being less decisive if the difference in value does not merit decisiveness.

When using SFA, learning features corresponding to one movement statistics while following another is not a new idea. \textcite{Franzius2007} proposed a general mechanism called \textit{learning rate adaptation} (LRA) to learn features encoding the orientation from movement in which the position changes quickly and vice versa. This works by up- or down-regulating learning for transitions in which the relative change in orientation is fast or slow. 
This can be applied to the setting in this work as well: If a transition between two states $s_u$ and $s_v$ has a high probability relative to other transitions, it is scaled down. If the transition has low probability, it is scaled up. This is called \textbf{LRA correction} in the following. Specifically, each transition is be scaled by the inverse of its transition probability $\frac{1}{P_{uv}}$ in the objective function. It is straightforward to confirm that this leads to a different objective from equation \eqref{eq:prob_sum_objective}:
\begin{align}
    \sum_{u, v} \mu_{u}P_{uv} \frac{1}{P_{uv}}(g_i(s_u) - g_i(s_v))^2 = 
    \sum_{u, v} \mu_{u}(g_i(s_u) - g_i(s_v))^2
\end{align}
where only non-zero transitions are considered. This has the consequence that for $M_{uv}=\frac{\mu_u + \mu_v}{2}$ for all pairs of states that have non-zero transition probability. 
Furthermore, for the diagonal matrix $\mathbf{D}$ the diagonal elements become $D_{vv} = \frac{1}{2}+\frac{N_v\mu_v}{2}$ where $N_v$ is the number of states connected by non-zero transition probabilities. The variance constraint does not change considerably, with only the least connected states contributing slightly less to the overall variance \footnote{Since spatial environments typically are similarly connected in terms of overall unweighted degree, this does not have a large overall effect.}. 
\begin{figure}[h]
    \centering
    \includegraphics[width=0.75\linewidth]{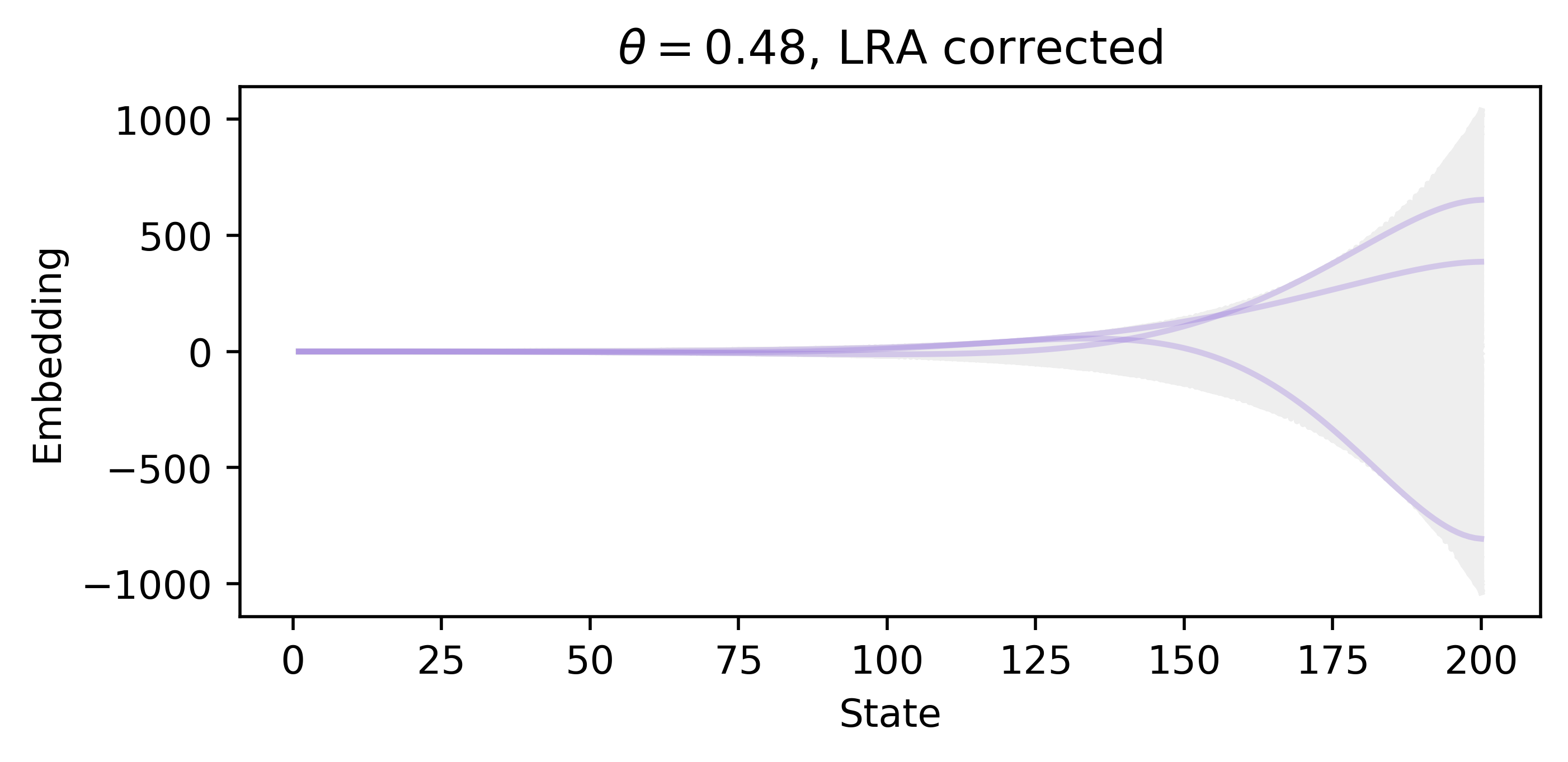}
    \vspace*{-4mm}
         \caption[Example of LRA on birth-death process]{Optimal features for the birth-death-process with $N=200$ and $\theta = 0.48$ after LRA correction. Best three embeddings shown in purple, all other features superimposed in gray.}
    \label{fig:bdprocess_features_lra_corrected}
\end{figure}
Figure \ref{fig:bdprocess_features_lra_corrected} illustrates the resulting features, which did not change considerably in this case.

\begin{figure}[h]
    \centering
    \includegraphics[width=0.75\linewidth]{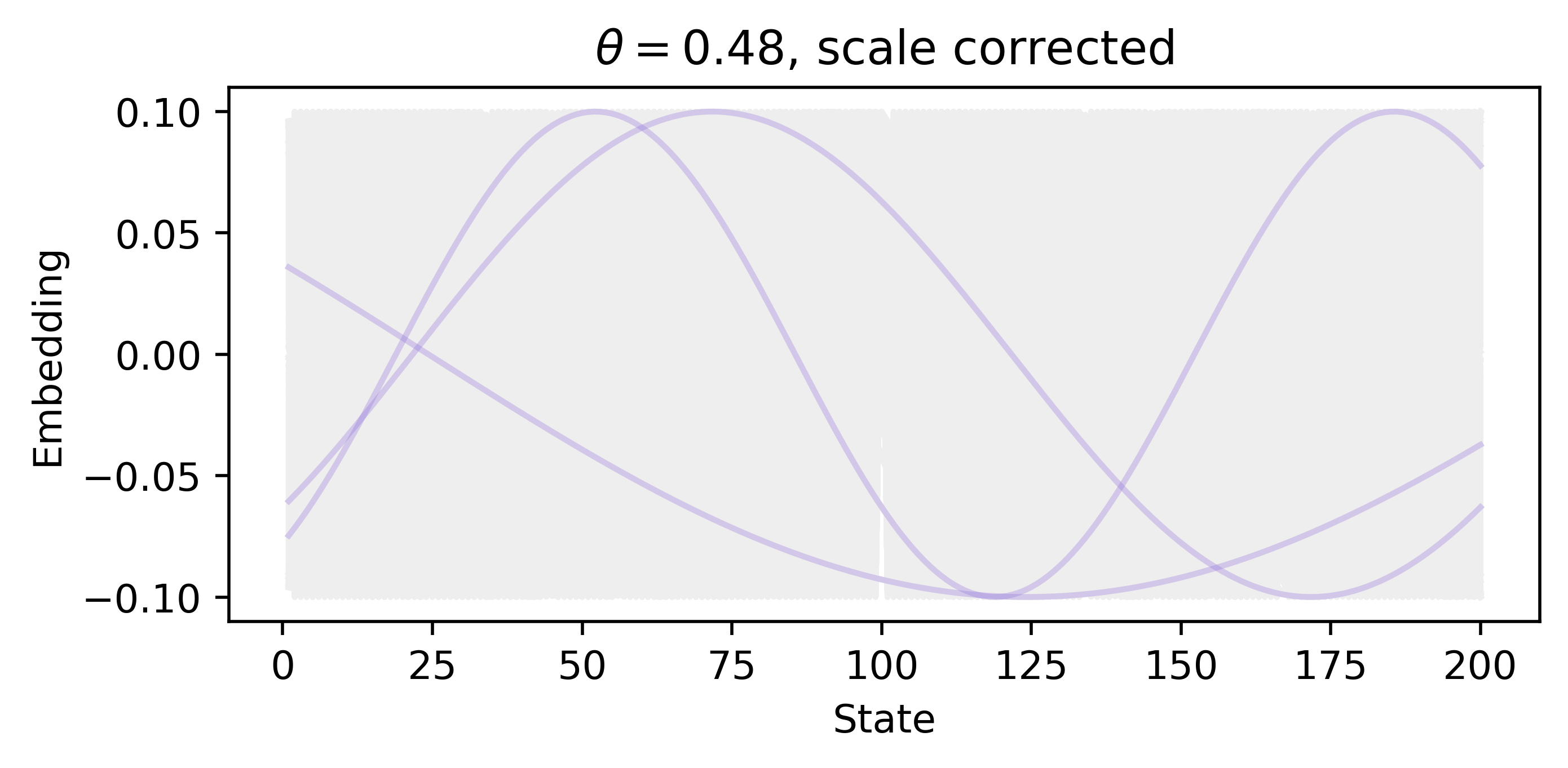}
    \vspace*{-4mm}
         \caption[Example of scale correction on birth-death process]{The optimal embeddings for the birth-death-process with $N=200$ and $\theta = 0.48$ after scale correction with $\mathbf{D}^\frac{1}{2}$. Best three embeddings shown in purple, all other features superimposed in gray.}
    \label{fig:bdprocess_features_repaired}
\end{figure}

A final correction mechanism that can be applied after sample collection and after feature extraction is a \textbf{scale correction}. The bounds in equation \eqref{eq:feature_bounds} indicate that the features of a point $i$ are scaled proportionally to $\frac{1}{\sqrt{\mu_i}}$. This implies that the feature of a point can be rescaled by multiplication with $\sqrt{\mu_i}$ or a full set of slow features $\mathbf{Y}$ can be rescaled by $\mathbf{D}^{\frac{1}{2}}\mathbf{Y}$. The result of such rescaling for the birth-death-process can be seen in Figure \ref{fig:bdprocess_features_repaired}.
The correction corresponds to moving the points from the feasible region of the constraint $\mathbf{Y}^T\mathbf{D}\mathbf{Y}=\mathbf{I}_{e+1}$ to the feasible region of $\mathbf{Y}^T\mathbf{Y}=\mathbf{I}_{e+1}$. Note that this will generally not result in the same features as extracted from a uniform policy, e.g., when considering zero-crossings and boundaries, but overall they exhibit uniform scaling (gray) and resemble uniform slow features. 

Although not used in this work, it is noteworthy that this correction can be generalized to any two variance constraints of the form \ref{eq:mcsfa_constraint_unitvariance_short}: $\mathbf{Y}^T\mathbf{\Omega}\mathbf{Y}=\mathbf{I}_{e+1}$ and $\mathbf{Y}^T\mathbf{\Phi}\mathbf{Y}=\mathbf{I}_{e+1}$, respectively, with diagonal matrices $\mathbf{\Omega}$ and $\mathbf{\Phi}$ with stationary distributions on the diagonal.

Their corresponding feasible regions $F_\mathbf{\Omega}$ and $F_\mathbf{\Phi}$ are related through a bijection $f$ as: 
\begin{align*}
f: F_\mathbf{\Omega} &\to F_\mathbf{\Phi}\\
\mathbf{y} &\mapsto \mathbf{\Phi}^{-\frac{1}{2}}\mathbf{\Omega}^{\frac{1}{2}}\mathbf{y}.
\end{align*}
since for $\mathbf{y}$ with $\mathbf{y}^T\mathbf{\Omega}\mathbf{y}=1$ 
\begin{align*}
    f(\mathbf{y})^T \mathbf{\Phi} f(\mathbf{y}) &= 
    (\mathbf{\Phi}^{-\frac{1}{2}}\mathbf{\Omega}^{\frac{1}{2}}\mathbf{y})^T \mathbf{\Phi} \mathbf{\Phi}^{-\frac{1}{2}}\mathbf{\Omega}^{\frac{1}{2}}\mathbf{y} \\
    &= 
    \mathbf{y}^T\mathbf{\Omega}^{\frac{1}{2}} \mathbf{\Phi}^{-\frac{1}{2}} \mathbf{\Phi} \mathbf{\Phi}^{-\frac{1}{2}}\mathbf{\Omega}^{\frac{1}{2}}\mathbf{y} \\
    &= 
    \mathbf{y}^T\mathbf{\Omega}^{\frac{1}{2}} \mathbf{\Omega}^{\frac{1}{2}}\mathbf{y} \\
    &=   \mathbf{y}^T\mathbf{\Omega}\mathbf{y} \\
    &= \mathbf{I}.  
\end{align*}
Validity and invertibility are consequences of $\mathbf{\Phi}$ and $\mathbf{\Omega}$ being diagonal matrices with strictly positive diagonal entries due to the ergodicity of the Markov chain.

Thus, $f$ and $f^{-1}$ correspond to coordinate-wise scaling of each $y_r$ by a factor $\sqrt{\frac{\omega_r}{\phi_r}}$ and $\sqrt{\frac{\phi_r}{\omega_r}}$, respectively.
If a state is more highly frequented under $\mathbf{\phi}$ than $\mathbf{\omega}$, this will lead to a systematic down-scaling of all features for this state, and the inverse holds true as well. 

\section{Experiments on Value Function Approximation using SFA}
\label{sec:mcsfa_approximation}
Although often inherently interpretable, the dominant role of slow features in machine learning is their use as basis for subsequent approximation on the input domain. As mentioned above, the target for approximation is the optimal value function $V^*(s)$, which is determined by standard dynamic programming \parencite{Sutton98} directly from the dynamics of the environment. 
 
Two settings are investigated, a linear graph environment similar to the birth-death process described in Section \ref{subsec:weakly} and a 2D lattice environment, with respect to the mean squared error of the resulting approximation. 

Stochasticity and directedness are induced solely by the policy, as is clarified in the corresponding sections. For both environments, $\zeta$-greedy behavior and Boltzmann behavior are evaluated for different degrees of goal-directedness and goal-aversion, each leading to different stationary distributions and thus different optimal slow features. These evaluations are repeated for different reward locations.

\subsection{Linear Graph Environment}
The linear graph environment used is similar to the birth-death process in Section \ref{subsec:weakly}, but instead of a single homogeneous transition probability $\theta$, a more general variant with individual $\theta_i$ for each state $s_i$ is used, which is defined solely by the behavior policy. Furthermore, the environment possesses a reward location $T$, so that $R(s_T) = 1$. This leads to:
\begin{alignat*}{3}
i\neq T:\quad& \theta_i &&= \pi(\text{right}|s_i)\\ 
i=T: \quad& \theta_i &&= 0.5
\end{alignat*}
where $\pi(\text{right}|s_i) + \pi(\text{left}|s_i) = 1$. In the following, such a process is called goal-directed, when 
\begin{alignat*}{3}
\forall i > T:\quad&\pi(\text{left}|s_i) &&> 0.5 \\
\forall i < T:\quad&\pi(\text{right}|s_i) &&> 0.5
\end{alignat*} and goal-averse, when
\begin{alignat*}{3}
\forall i > T:\quad&\pi(\text{left}|s_i) && < 0.5 \\
\forall i < T:\quad&\pi(\text{right}|s_i) &&> 0.5
\end{alignat*} 
meaning that the process will move towards or away from the goal-location with higher probability, respectively. A process in which all actions are equally likely for all states is called uniform.

\begin{figure}[h]
    \centering
    \includegraphics[width=0.95\linewidth]{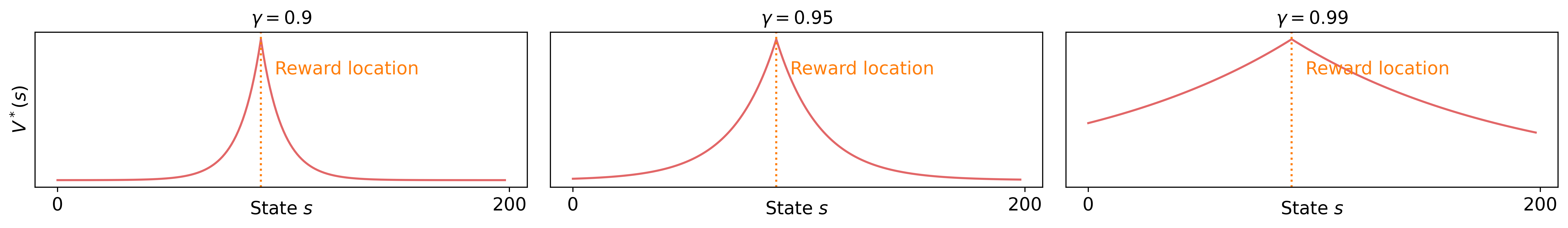}
    \caption{Value function of optimal policy for different $\gamma$ on a linear graph with 200 states and goal-location 90.}
    \label{fig:mcsfa_optimal_value_function_example_1d}
\end{figure}

Figure \ref{fig:mcsfa_optimal_value_function_example_1d} illustrates the value function $V^*(s)$ for 200 states and a particular reward location for different discount factors. For all subsequent investigations, $\gamma=0.95$ is used, but the results qualitatively transfer to different discount factors because the general shape is not affected.

\paragraph{$\zeta$-greedy behavior} First, the case of a $\zeta$-greedy policy is considered, where in each state, the action that leads an agent away from the goal's location has probability $\zeta \in [0,1]$ independent of the actual state. This behavior induces a goal-directed process for $\zeta < 0.5$, a uniform process for $\zeta = 0.5$, and a goal-averse process for $\zeta > 0.5$.

\begin{figure}[h]
    \centering
    \includegraphics[width=0.99\linewidth]{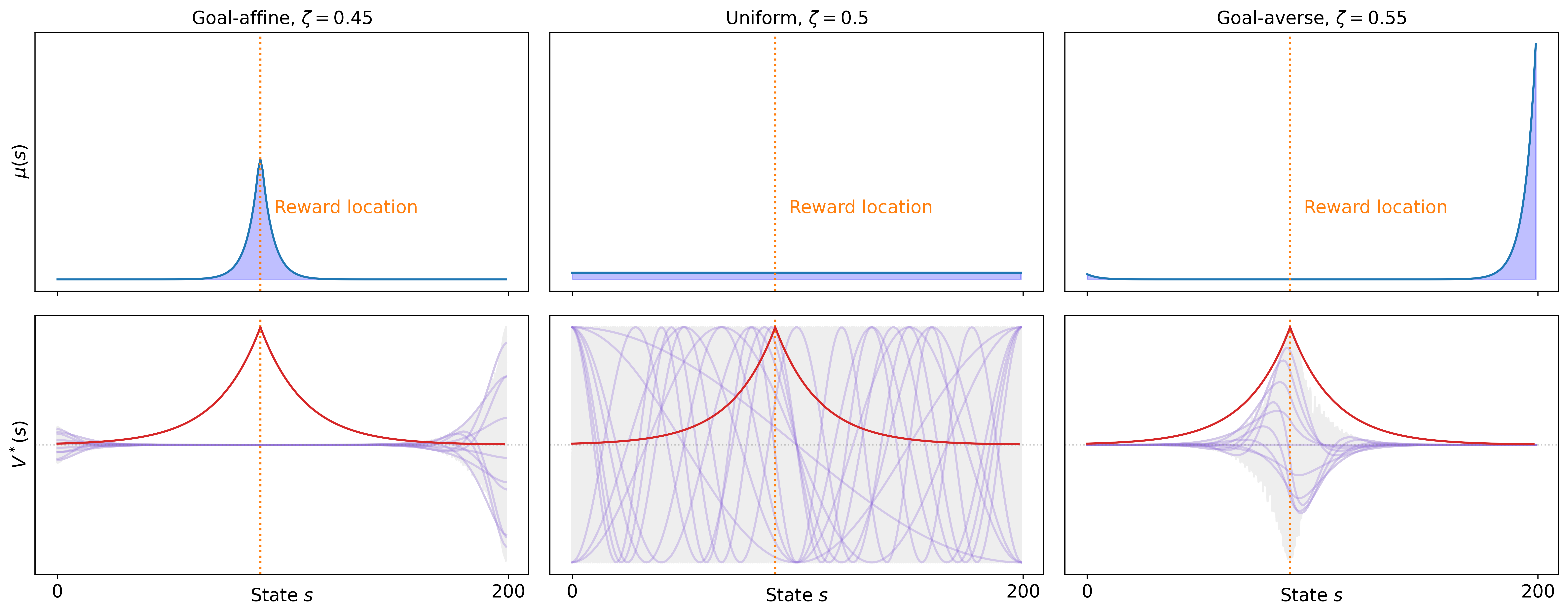}
    \caption[Stationary distribution, optimal slow features and optimal value function on a linear graph following $\zeta$-greedy behavior.]{Illustration of the effect of $\zeta$-greedy behavior on the stationary distribution and slow features of the birth-death-process. \textbf{Top:} Stationary distributions. \textbf{Bottom:} Optimal value function for $\gamma=0.95$ and overlay of the first ten slow features of the Markov chain. All features superimposed in gray.}
    \label{fig:mcsfa_zeta_averse_and_directed_behavior_example_1d}
\end{figure}

The first row of Figure \ref{fig:mcsfa_zeta_averse_and_directed_behavior_example_1d} shows the resulting stationary distributions for different $\zeta$ for an example environment. The second row shows the first slow features (purple) and the whole set (gray) of the induced Markov chain and the optimal value function. Confirming the findings of Section \ref{subsec:weakly}, the overall feature scale is flat in the region of most occupancy, which coincides with the reward location for goal-directed behavior. For uniform behavior, they are unrelated and the scale is uniform. For goal-averse behavior, feature scaling and shape of value-function coincide. 

\begin{figure}[h]
    \centering
    \includegraphics[width=0.99\linewidth]{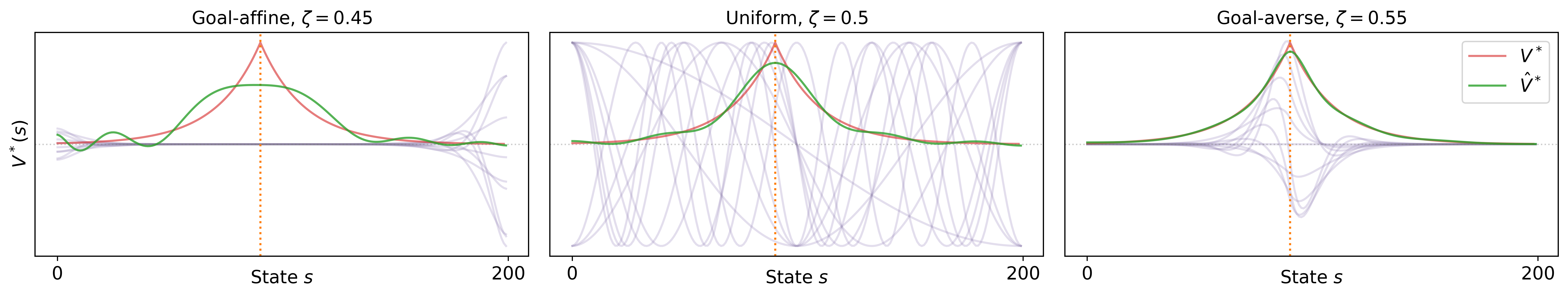}
    \caption{Illustration of the possible effects of $\zeta$-greedy behavior on the quality of approximation.}
    \label{fig:mcsfa_zeta_regression_example_1d}
\end{figure}

The consequence of this for regression is illustrated in Figure \ref{fig:mcsfa_zeta_regression_example_1d}. When comparing the value function with the best ordinary least squares approximation using the first ten features, a possible detrimental effect of slow features from goal-directed behavior becomes clear: The value function naturally peaks around the reward location, while goal-directed behavior leads to flattened features at exactly this position leading to a flattened approximation. A potentially beneficial effect can be seen for goal-averse behavior.

\begin{figure}[h]
    \centering
    \includegraphics[width=0.99\linewidth]{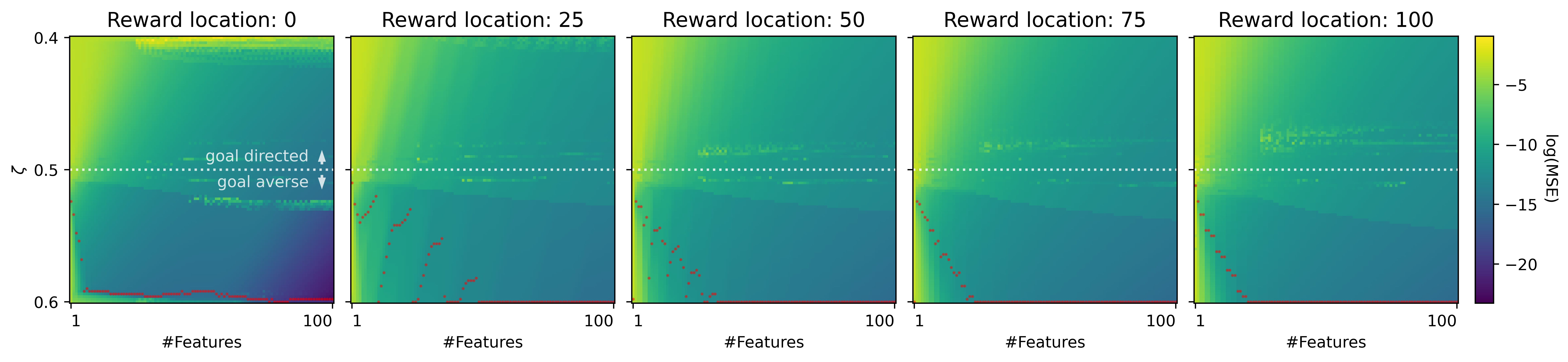}
    \caption{The regression performance as log mean-squared-error for different reward positions, dimension of embedding, and goal-affinities when using $\zeta$-greedy behavior to induce the Markov chain. Red dots indicates best performance for each number of features.}
    \label{fig:mcsfa_bdprocess_approximation_zeta_results}
\end{figure}

\noindent To further investigate the effect on approximation quality, experiments were conducted for different reward positions on the left half of the state space. Corresponding positions in the right half will result in mirrored results due to the symmetry of behavior and MCSFA. For each position, the logarithm of the mean-squared-error is reported depending on the number of features (the dimension of the embedding) and varying degrees of goal-directedness $\zeta$. Figure \ref{fig:mcsfa_bdprocess_approximation_zeta_results} shows the results.

The approximation performance is reduced when using features based on goal-directed behavior compared to uniform or goal-averse behavior, indicating that slowness optimization in the SFA-sense and goal-driven behavior in spatial environments are potentially misaligned objectives in a reinforcement learning setting. This is aggravated by the fact that the discrepancy is most pronounced when a low embedding dimensionality is used, which is an aim of representation learning and dimensionality reduction in general and slow feature analysis in particular. Behaving increasingly goal-directed or -averse beyond the tested values can lead to states with extremely low occupancy, due to its exponential nature, and thus numerical instability, as visible at the borders and thus no stronger directedness or aversion is considered. 

In the following, the corrections proposed in Section \ref{sec:mcsfa_correction_mechanisms} are evaluated for their influence on approximation quality.

\begin{figure}[h!]
    \centering
    \includegraphics[width=0.99\linewidth]{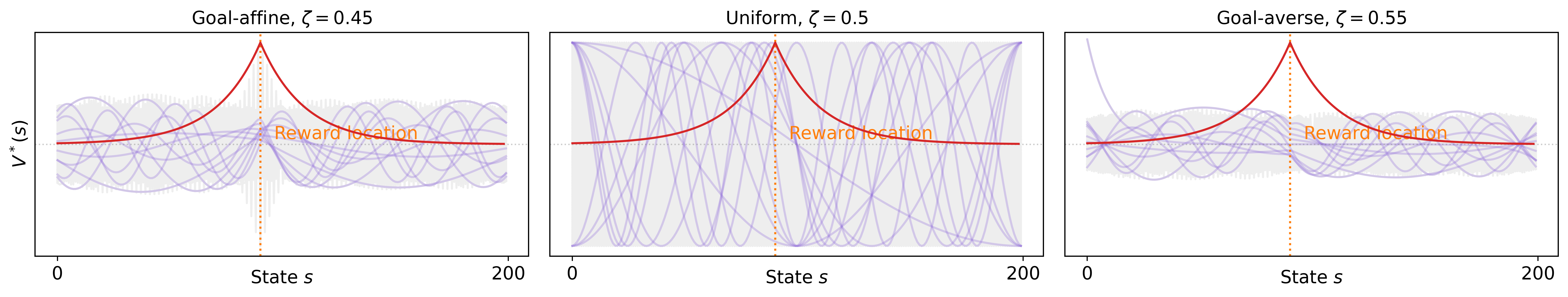}
    \caption{Illustration of the effect of scale correction on features from $\zeta$-greedy behavior.}
    \label{fig:mcsfa_features_zeta_example_1d_repair}
\end{figure}

\paragraph{Scale correction}
The effect of rescaling the features is illustrated in Figure \ref{fig:mcsfa_features_zeta_example_1d_repair} for examples of goal-directed, uniform, and goal-averse behavior.

\begin{figure}[h]
    \centering
    \includegraphics[width=0.99\linewidth]{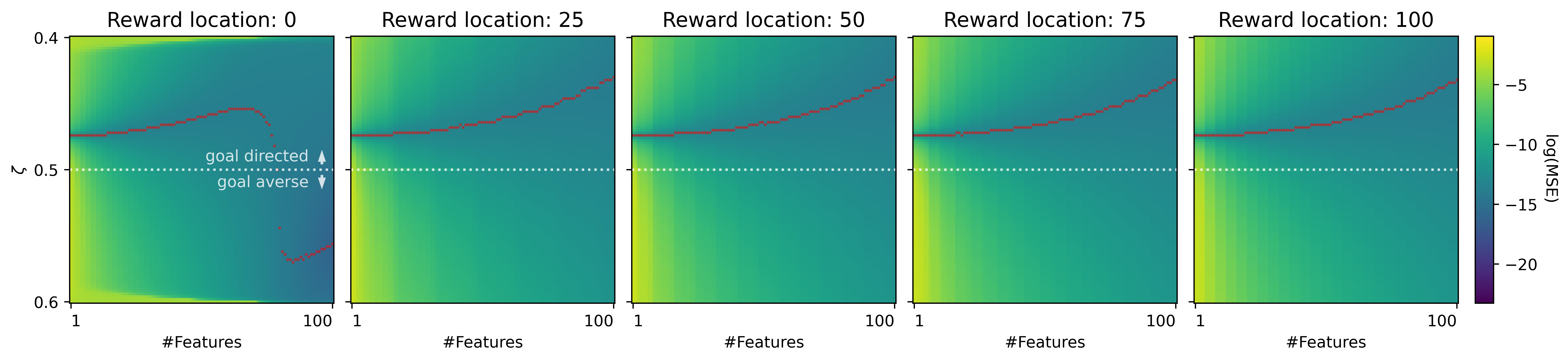}
    \caption{The regression performance as log mean squared error for different reward positions, dimension of embedding, and goal-affinities when using $\zeta$-greedy behavior to induce the Markov chain after applying scale correction to features. Red dots indicates best performance for each number of features.}
    \label{fig:mcsfa_experiments_zeta_1d_repair}
\end{figure}

The extreme scaling is counteracted to some extent, but the features still possess characteristics different from those acquired from the uniform behavior. This accounts for the differences in the regression performances, which are shown in Figure \ref{fig:mcsfa_experiments_zeta_1d_repair} with the same color scale as used in Figure \ref{fig:mcsfa_bdprocess_approximation_zeta_results}. However, it is apparent that after rescaling, performance is significantly less influenced by overall behavior. 

\begin{figure}[h]
    \centering
    \includegraphics[width=0.99\linewidth]{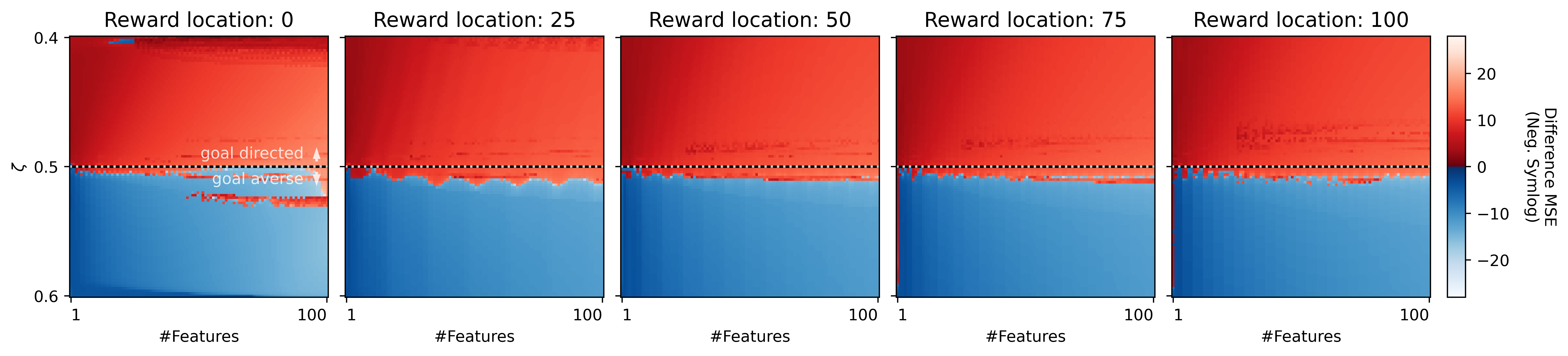}
    \caption{The difference in regression performance as symlog mean-squared-error after the scale correction. Plotted for different reward positions, dimension of embedding, and goal-affinities when using $\zeta$-greedy behavior to induce the Markov chain. Negative values (blue) indicate decreased performance, positive values (red) indicate increased performance. Deeper saturation indicates stronger effect size.}
    \label{fig:mcsfa_experiments_diff_zeta_1d_repair}
\end{figure}

In fact, rescaled goal-directed features seem to lead to better approximation performances when compared to rescaled goal-averse features. However, this effect is actually caused by impeding the performance of goal-averse features. This becomes clear when comparing the performance before and after scaling in Figure \ref{fig:mcsfa_experiments_diff_zeta_1d_repair}. The reported metric is $-\symlog(\text{MSE}_\text{original} - \text{MSE}_\text{corrected})$ to report the difference, where 
\begin{equation}
    \symlog(x) = 
\begin{cases}
    \sgn(x) \cdot \log |x|,\quad x \neq 0\\
    0,\quad \text{else}.
\end{cases}
\end{equation}

Positive values (red) indicate an improvement and negative values (blue) indicate a worsening of performance.
For all reward locations, the correction has a largely detrimental effect when used on goal-averse features and an exclusively beneficial effect on goal-directed features in the one-dimensional setting. This emphasizes the hypothesis that goal-averse scaling can be beneficial to approximation performance, and, in this setting, correcting for it will reduce performance. At the same time, goal-directed scaling negatively impacts performance, leading to a beneficial effect of the correction.

\begin{figure}[h]
    \includegraphics[width=1\linewidth]{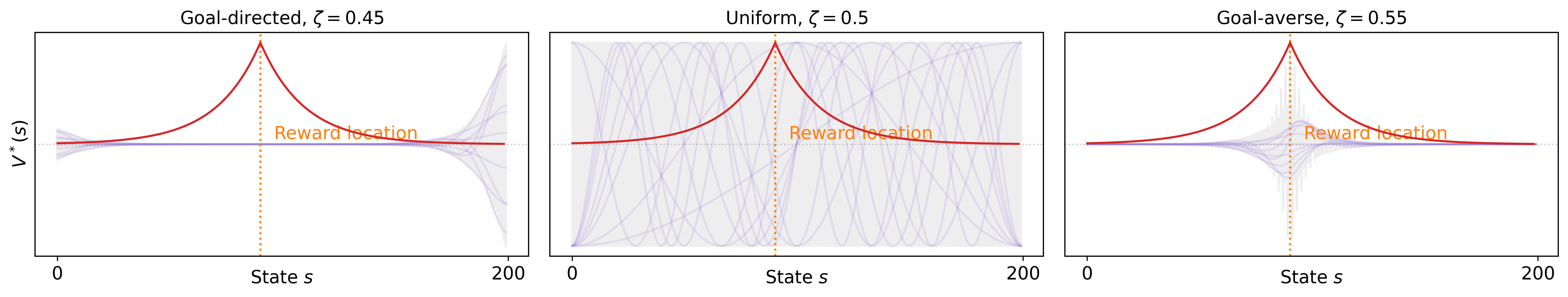}
    \caption{Example illustration of the effect of LRA correction on features from $\zeta$-greedy behavior.}
    \label{fig:mcsfa_experiments_zeta_1d_constraintrepair_illustration}
\end{figure}

\begin{figure}[h!]
    \centering
    \begin{subfigure}{0.98\textwidth}
        \includegraphics[width=1\linewidth]{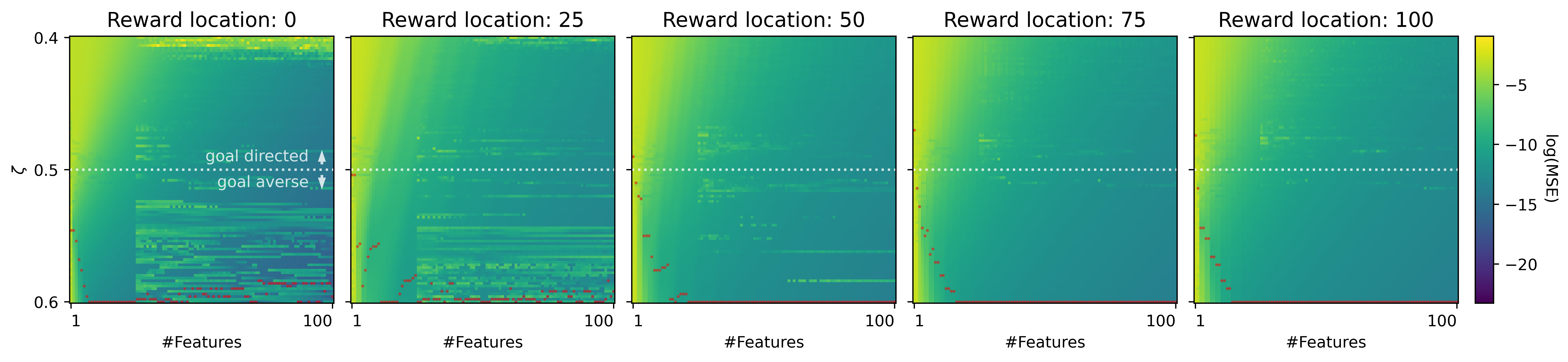}
        \caption{}
        \label{fig:mcsfa_experiments_zeta_1d_constraintrepair_performance}
    \end{subfigure}
    \vspace{0.3cm}
    \begin{subfigure}{0.99\textwidth}
        \includegraphics[width=1\linewidth]{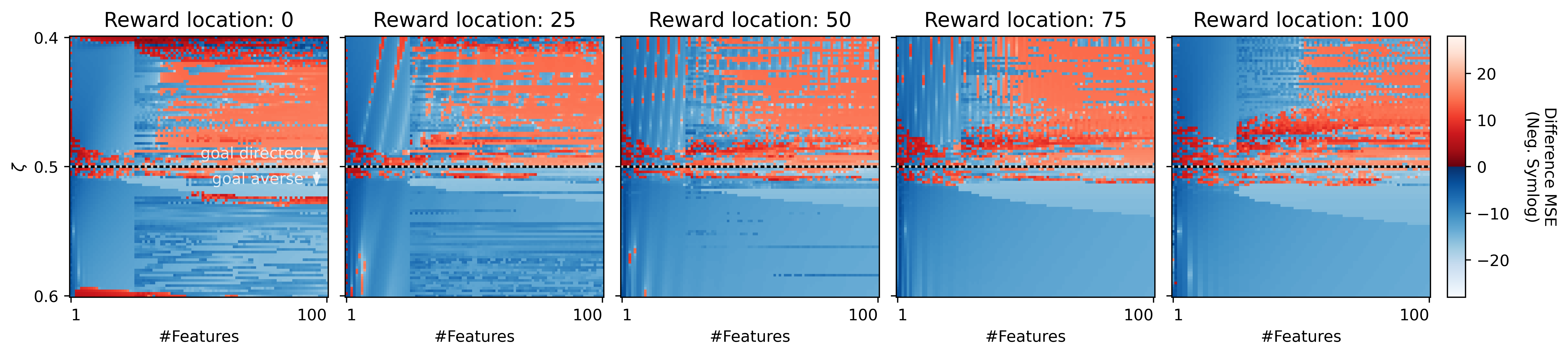}
        \caption{}
        \label{fig:mcsfa_experiments_zeta_1d_constraintrepair_diff}
    \end{subfigure}
    \caption{(a) The regression performance for different reward positions, dimension of embedding, and goal-affinities after applying LRA correction to $\zeta$-greedy behavior. Red dots indicates best performance for each number of features. (c) The difference in regression performance after the LRA correction. Negative values (blue) indicate decreased performance, positive values (red) indicate increased performance. Deeper saturation indicates stronger effect size.}
    \label{fig:mcsfa_experiments_zeta_1d_constraintrepair}
\end{figure}

\paragraph{LRA correction} 
The qualitative effect of the LRA correction on the features, the approximation performance, as well as the difference to uncorrected $\zeta$-greedy behavior are displayed in Figures \ref{fig:mcsfa_experiments_zeta_1d_constraintrepair_illustration} and \ref{fig:mcsfa_experiments_zeta_1d_constraintrepair}. The resulting procedure is less numerically stable, but an overall trend is recognizable.

Although the correction has a mild effect on feature scaling, a positive effect is recognizable for some settings of goal-directed behavior, particularly with an increased number of features, as can be seen in Figure  \ref{fig:mcsfa_experiments_zeta_1d_constraintrepair_diff}. 
For goal-averse $\zeta$-greedy behavior, the LRA correction is largely detrimental to theapproximation performance. As a result, as with scale correction, the approximation performance generally exhibits less dependency on the behavior, although the best performances are still achieved in the most goal-averse settings.

\begin{figure}[h]
    \centering
    \includegraphics[width=0.99\linewidth]{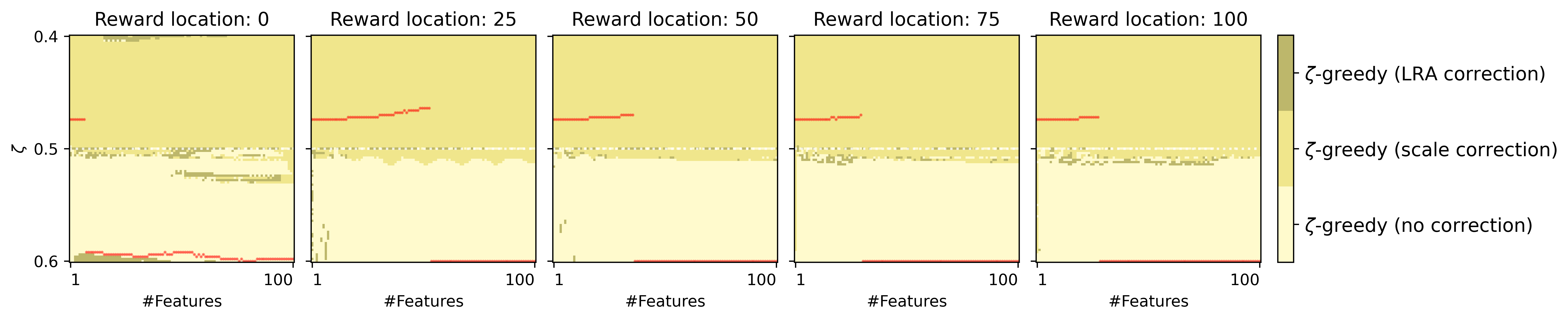}
    \caption{Best approximation performances for $\zeta$-greedy behavior for different settings and corrections. Best performance per feature dimension indicated in red.}
    \label{fig:mcsfa_compare_zeta_1d}
\end{figure}

Figure \ref{fig:mcsfa_compare_zeta_1d} compares all three variants of $\zeta$-greedy features by choosing the best performance for each configuration. 
In alignment with the intuition and results discussed above, this implies that scale correction is largely beneficial in the setting of goal-directed behavior, while for goal-averse behavior the uncorrected features perform best, possibly due to coinciding scaling with the value-function. This is also reflected in the best performances overall being achieved largely by performing most goal-averse. Except for artifacts, LRA corrected behavior does not seem to be beneficial in the linear graph environment.

\begin{figure}[h!]
    \centering
    \includegraphics[width=0.95\linewidth]{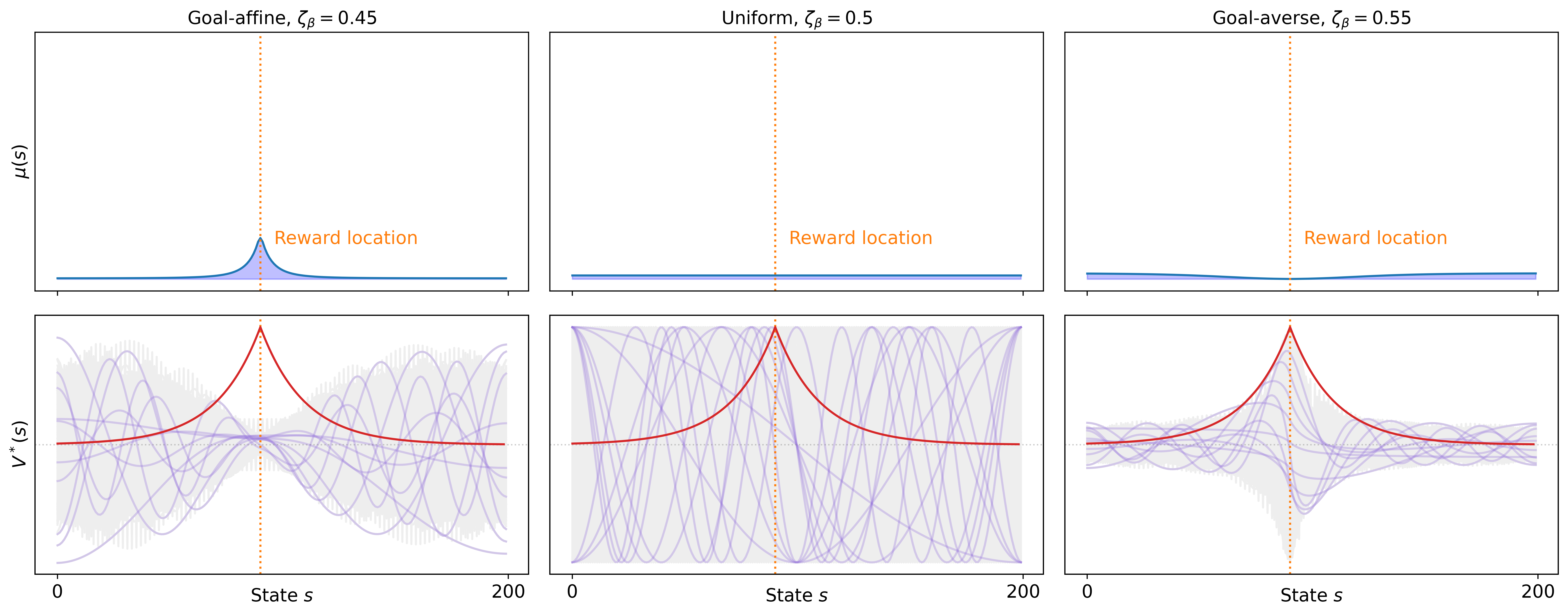}
    \caption{Illustration of the effect of Boltzmann behavior on the stationary distribution and slow features of the birth-death-process. \textbf{Top:} Stationary distributions. \textbf{Bottom:} Optimal value function for $\gamma=0.95$ and overlay of the first ten slow features of the Markov chain. All features superimposed in gray.}
    \label{fig:mcsfa_boltzmann_averse_and_directed_behavior_example_1d}
\end{figure}

\begin{figure}[h]
    \centering
    \includegraphics[width=0.99\linewidth]{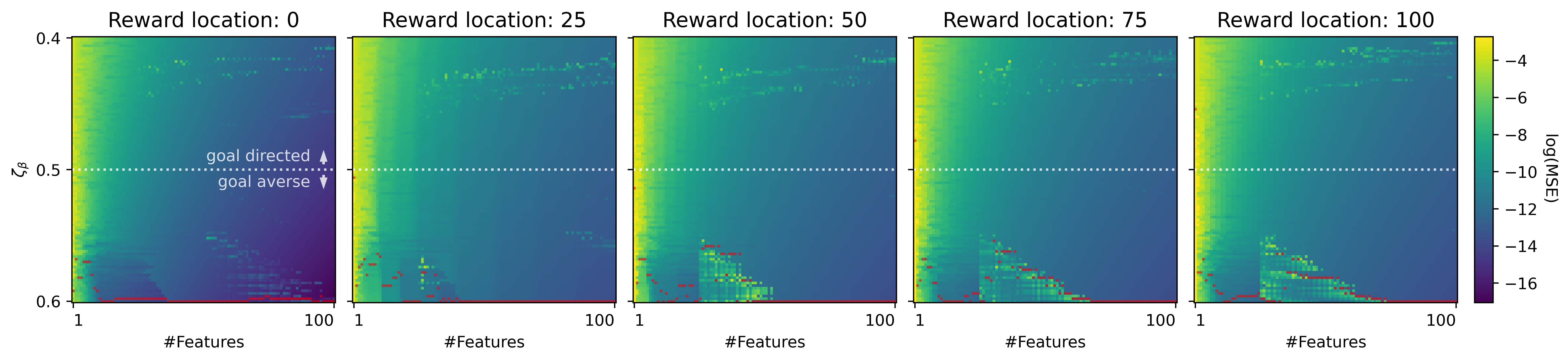}
    \caption{The approximation quality as log mean-squared-error for different reward positions, dimension of embedding, and goal-affinities when using Boltzmann behavior to induce the Markov chain.}
    \label{fig:mcsfa_bdprocess_approximation_boltzmann_results}
\end{figure}

\begin{figure}[h!]
    \centering
    \includegraphics[width=0.99\linewidth]{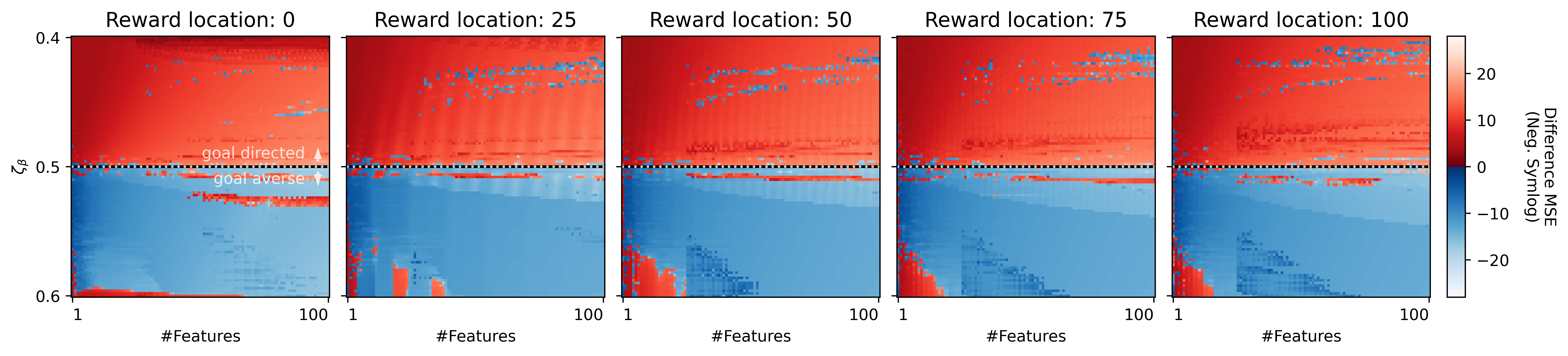}
    \caption{The difference of quality as symlog mean-squared-error when switching from $\zeta$-greedy to Boltzmann behavior for different reward positions, dimension of embedding, and goal-affinities. Negative values (blue) indicate decreased performance, positive values (red) indicate increased performance. Saturation indicates size of the effect.}
    \label{fig:mcsfa_experiments_diff_boltzmann_1d_vs_zeta}
\end{figure}

\paragraph{Boltzmann behavior}
For comparability between the $\zeta$-greedy behavior and Boltzmann exploration, $\beta$ was chosen to correspond to a given $\zeta_\beta$, such that states directly neighboring the goal have the same probability for selecting one of the optimal actions under the different policies and, consequently, an agent close to the reward location behaves similarly to $\zeta$-greedy behavior and becomes less decisive the farther away the reward location is. Thus, the occupancy is more evenly distributed, leading to less extreme scaling in the slow features as displayed in Figure \ref{fig:mcsfa_boltzmann_averse_and_directed_behavior_example_1d}.
However, in the approximation performance, the effect of the behavior is still pronounced (Figure \ref{fig:mcsfa_bdprocess_approximation_boltzmann_results}) as goal-averse features still significantly outperform goal-directed features for the approximation of $V^*$.
When directly comparing $\zeta$-greedy and Boltzmann behavior in Figure \ref{fig:mcsfa_experiments_diff_boltzmann_1d_vs_zeta}, it appears that the latter helps alleviate the detrimental effect of goal-directedness but performs slightly worse in the goal-averse case.

\begin{figure}[h!]
    \centering
    \includegraphics[width=0.99\linewidth]{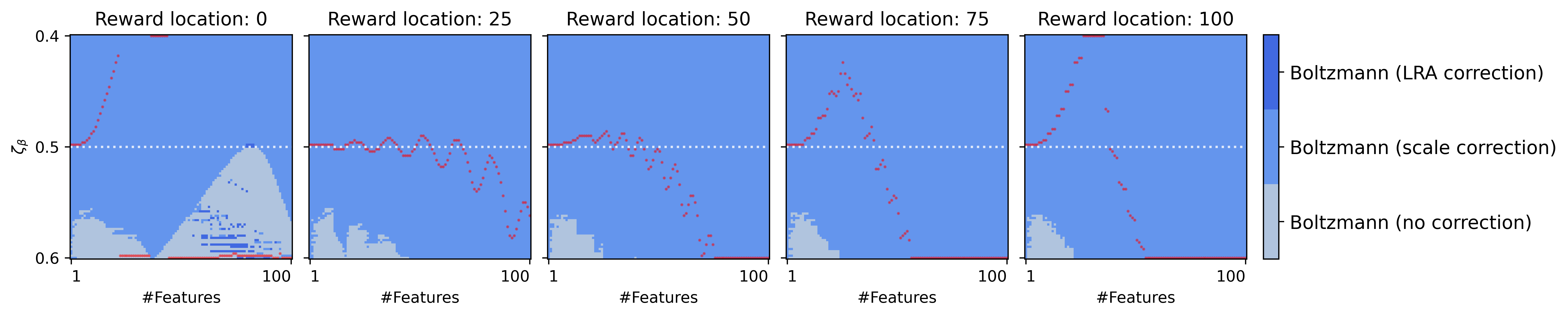}
    \caption{Best approximation performances for Boltzmann behavior for different settings and corrections. Best performance per feature dimension indicated in red.}
    \label{fig:mcsfa_compare_boltzmann_1d}
\end{figure}

Since the scale of features from Boltzmann behavior is still affected by the stationary distribution, the scaling and LRA correction mechanisms previously discussed can also be applied to it. Figure \ref{fig:mcsfa_compare_boltzmann_1d} compares the different variants of corrected or uncorrected features and indicates for which combination of feature dimension and behavior the best approximation performance is obtained.
For Boltzmann behavior in the linear graph environment, applying the scale correction results in overall better performance than uncorrected and LRA correction does not result in better performance in any setting. The best performance varies, but tends to goal-averse behavior when the number of features is increased.

\begin{figure}[h!]
    \centering
    \includegraphics[width=0.99\linewidth]{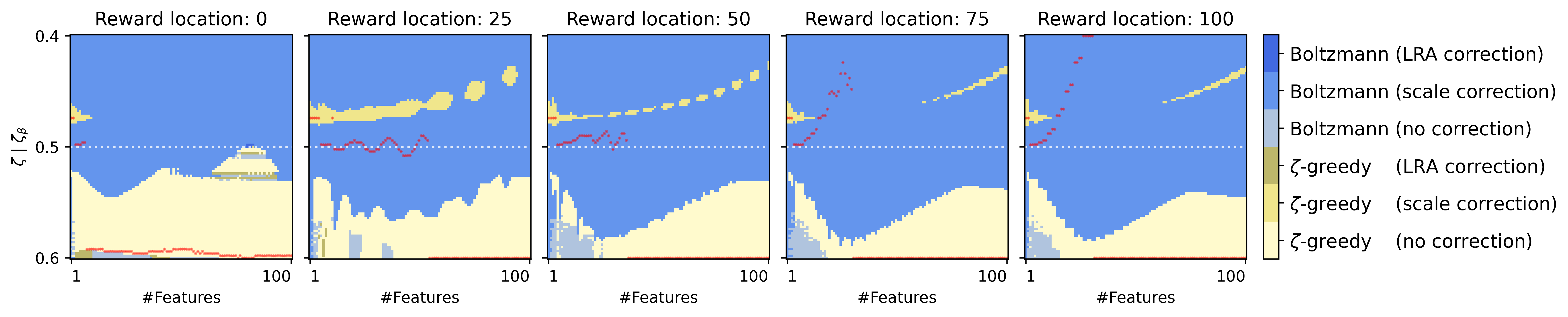}
    \caption{Best approximation performances for $\zeta$-greedy and Boltzmann behavior for different settings and corrections. Best performance per feature dimension indicated in red.}
    \label{fig:mcsfa_compare_zeta_1d_boltzmann_1d}
\end{figure}

\paragraph{Summary for linear graph environment} When comparing all discussed settings on the linear graph environment in Figure \ref{fig:mcsfa_compare_zeta_1d_boltzmann_1d}, one can conclude that Boltzmann behavior with scale correction is largely beneficial in the goal-directed and in the slightly goal-averse setting. In the more goal-averse settings, uncorrected $\zeta$-greedy behavior results in the best performance. 
LRA correction seems to be largely ineffective in this environment and, regardless of the choice of behavior or correction, goal-aversion in most cases yields the best features for the approximation of $V^*$ once a certain number of features is used for approximation. These results seem to confirm the intuition that it is beneficial when the scale of the features coincides with the function to be approximated.

\clearpage
\subsection{Lattice Graph Environment}
In this section, the same experiments are repeated for a lattice graph with $20\times20$ states\footnote{Size chosen according to available compute.}, organized in the fashion depicted as an example in Figure \ref{fig:single_room_graph}, \input{room_graph} with the possible actions $\uparrow,\, \downarrow,\,\leftarrow,\,\rightarrow$ leading to transitions into the corresponding states. On the sides, if no target node is available in the chosen direction, a self-transition will occur. 
Thus, the graph is a generalization of the previously discussed linear graph, and the notions of goal-directedness or goal-aversion can be naturally applied with the modification that there might be more than one optimal action. In these cases, greedy behavior is defined as assigning equal (goal-directed) probability to all optimal actions. Furthermore, any behavior that assigns non-zero probability to all actions in all states will result in an ergodic Markov chain. 

\begin{figure}[h]
    \centering
    \includegraphics[width=0.90\linewidth]{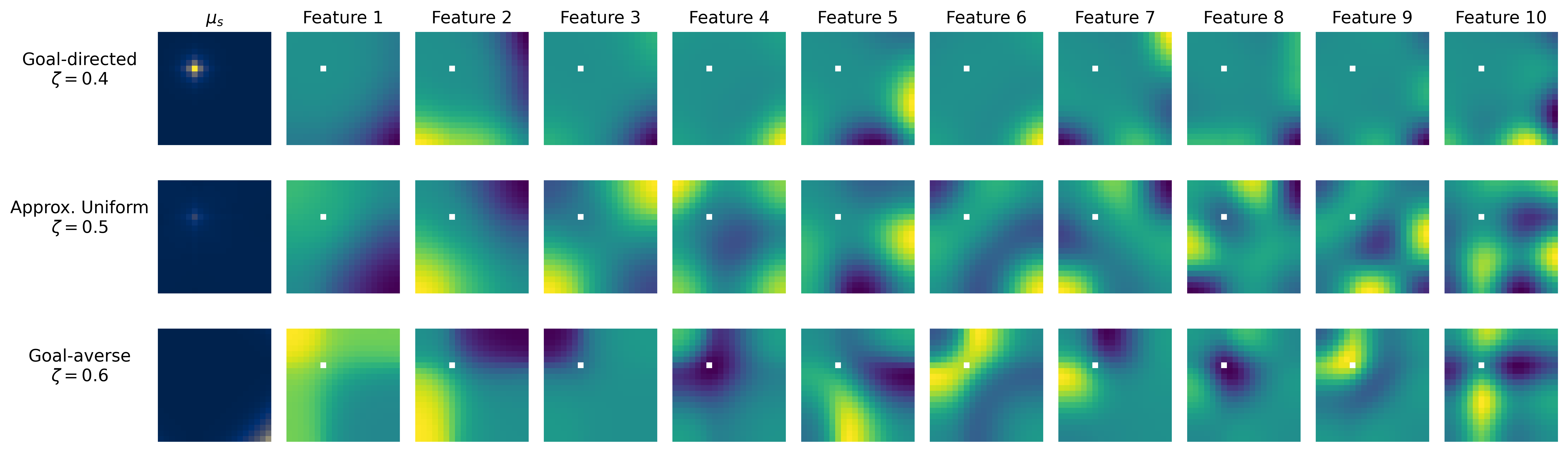}
    \caption{Stationary distribution and example features of $\zeta$-greedy behavior for different degrees of goal-directedness or aversion in a 2D lattice environment.}
    \label{fig:mcsfa_zeta_2d_lattice_features}
\end{figure}

Following the previous section, $\zeta$-greedy behavior is first investigated, with Figure \ref{fig:mcsfa_zeta_2d_lattice_features} showing illustrative examples of the resulting features.

The scaling effect is also present in the 2D lattice -- goal-directed behavior leads to flat regions around the goal position (indicated in white), while goal-averse behavior leads to the reverse. Note that for $\zeta=0.5$ they are not fully uniform due to the particular choice of policy parameterization and the fact that in the horizontal or vertical direction, there is only one optimal action, but this effect is accepted for the benefit of a continuous transition from goal-directed to goal-averse behavior.

\begin{figure}
    \centering
    \includegraphics[width=0.77\linewidth]{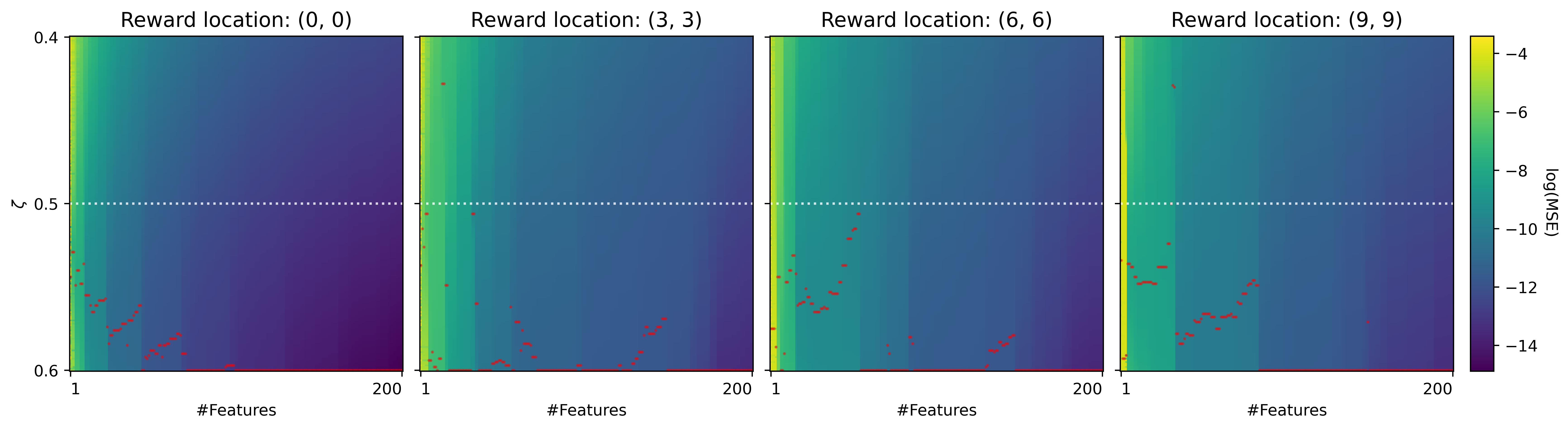}
    \caption{The regression performance as log mean-squared-error for different reward positions, dimension of embedding, and goal-affinities when using $\zeta$-greedy behavior to induce the Markov chain. Red dots indicates best performance for each number of features.}
    \label{fig:mcsfa_experiments_zeta_2d_standard}
\end{figure}

When comparing the regression results (Figure \ref{fig:mcsfa_experiments_zeta_2d_standard}) for different reward locations (in this case, two dimensional with (0,0) denoting the bottom-left corner) and $\zeta$-greedy behavior, one sees a similar effect as in the 1D case: Goal-averse behavior tends to reliably produce better regression results, although the effect is not as pronounced. As this scaling effect is generally present, this might be caused by the smaller maximal graph distance due to the construction of the environment, which in turn leads to even the low occupancy states being visited more often. The size is limited by computational considerations as the state space grows quadratic in the width / height of the environment, and the decomposed matrices grow quadratically in that state space.

\begin{figure}[h]
    \centering
    \includegraphics[width=0.9\linewidth]{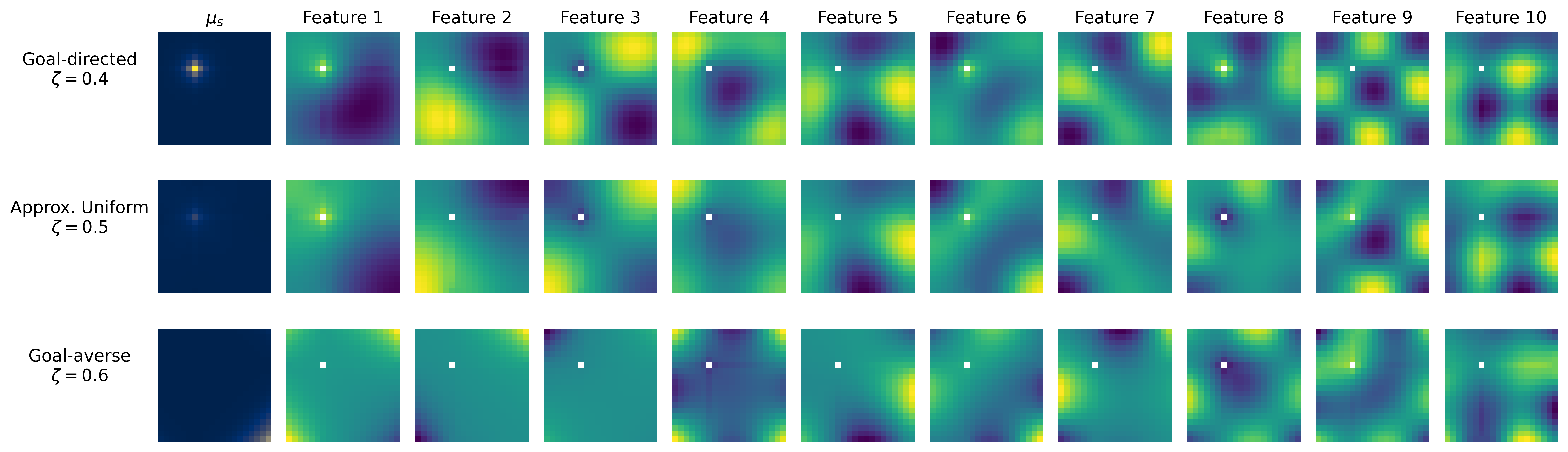}
    \caption{Stationary distribution and example features of $\zeta$-greedy behavior with scale correction applied for different degrees of goal-directedness or aversion in a 2D lattice environment.}
    \label{fig:mcsfa_zeta_2d_lattice_features_repair}
\end{figure}

\paragraph{Scale correction}
When scale correction is applied to the resulting features, as illustrated in Figure \ref{fig:mcsfa_zeta_2d_lattice_features_repair}, they exhibit a more uniform scaling. However, it appears it appears that the first goal-averse features are overcorrected to some extent.

\begin{figure}[t]
    \centering
    \begin{subfigure}{0.77\textwidth}
        \includegraphics[width=1\linewidth]{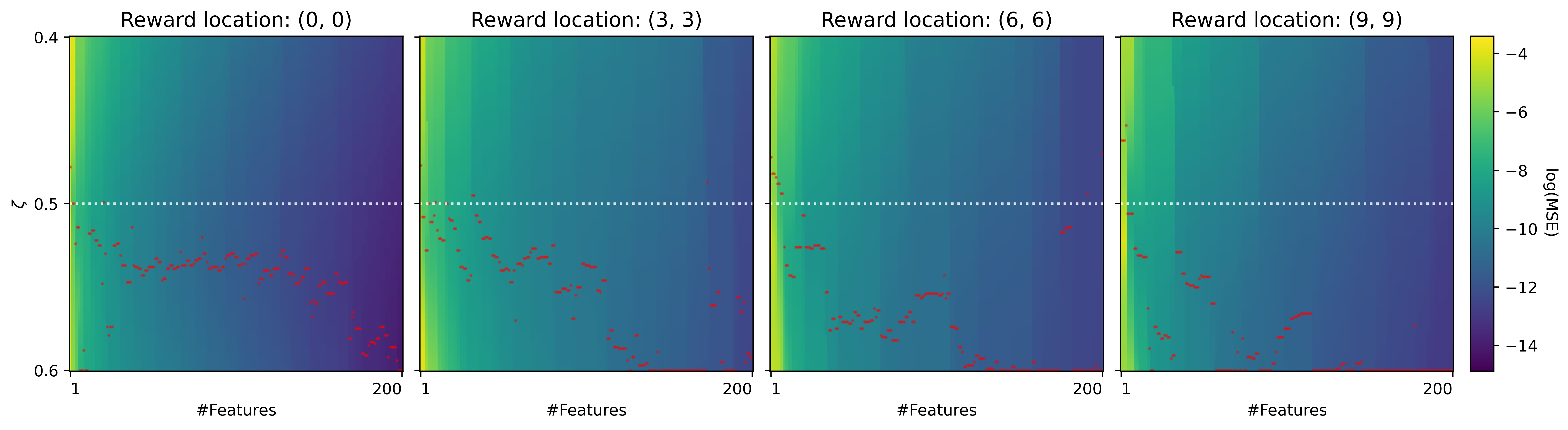}
        \caption{}
    \end{subfigure}
    \vspace{0.4cm}
    \begin{subfigure}{0.77\textwidth}
        \includegraphics[width=1\linewidth]{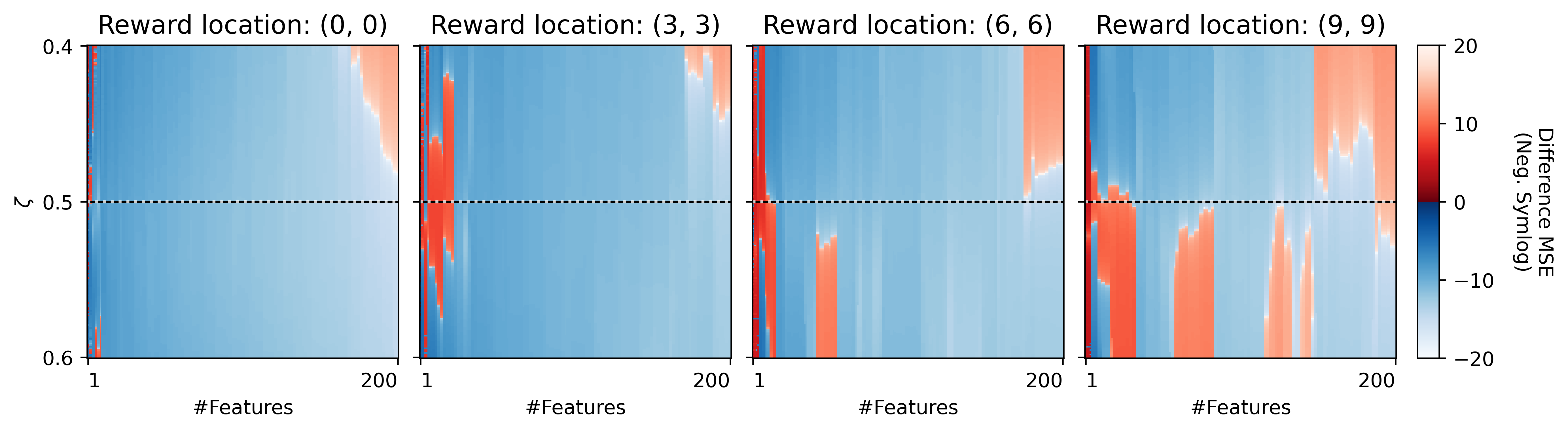}
        \caption{}
    \end{subfigure}
    \caption{Regression results in the 2D environment for $\zeta$-greedy behavior after applying feature scale correction (a) with difference visualized in (b).}
    \label{fig:mcsfa_experiments_zeta_2d_repair}
\end{figure}

The approximation performances are visualized in Figure \ref{fig:mcsfa_experiments_zeta_2d_repair}. Overall, the scale correction does not exhibit a large effect and, except for some settings, seems to be detrimental to overall performance. As before, the best performances for each number of features are achieved by goal-averse behavior.

\begin{figure}[h]
    \centering
    \includegraphics[width=0.9\linewidth]{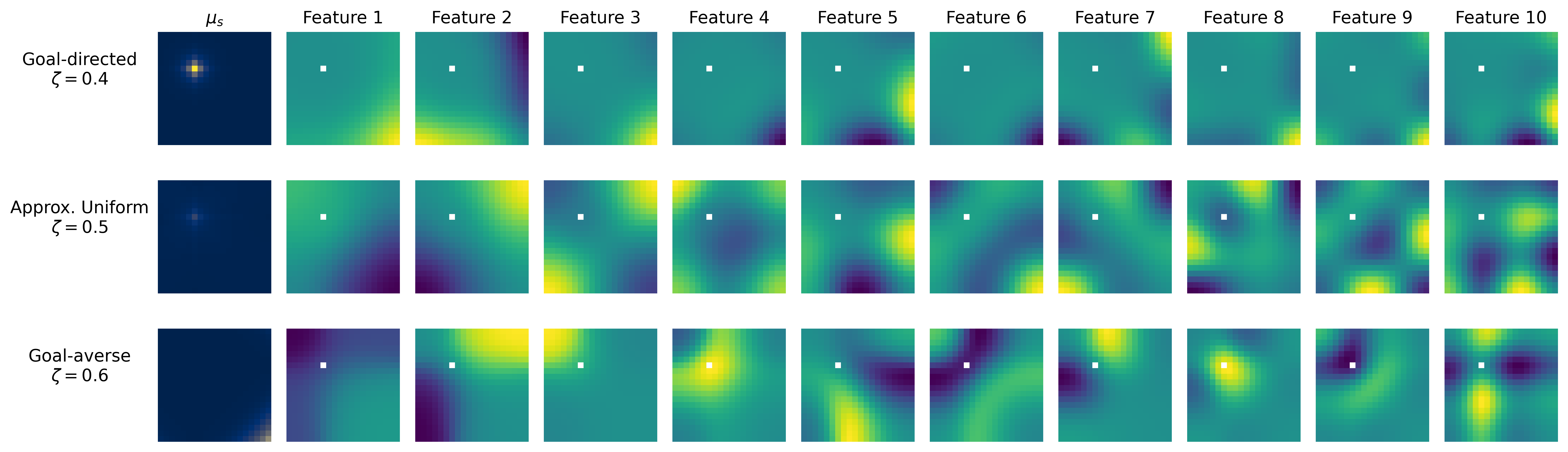}
    \caption{Stationary distribution and example features of $\zeta$-greedy behavior with LRA correction applied for different degrees of goal-directedness or aversion in a 2D lattice environment.}
    \label{fig:mcsfa_zeta_2d_lattice_features_constraintrepair}
\end{figure}

\newpage
\paragraph{LRA correction} 
Figure \ref{fig:mcsfa_zeta_2d_lattice_features_constraintrepair} shows the features resulting from the application of LRA correction to the optimization problem. It seems to be largely ineffective in correcting the scale of individual features.

\begin{figure}[ht]
    \centering
    \begin{subfigure}{0.77\textwidth}
        \includegraphics[width=1\linewidth]{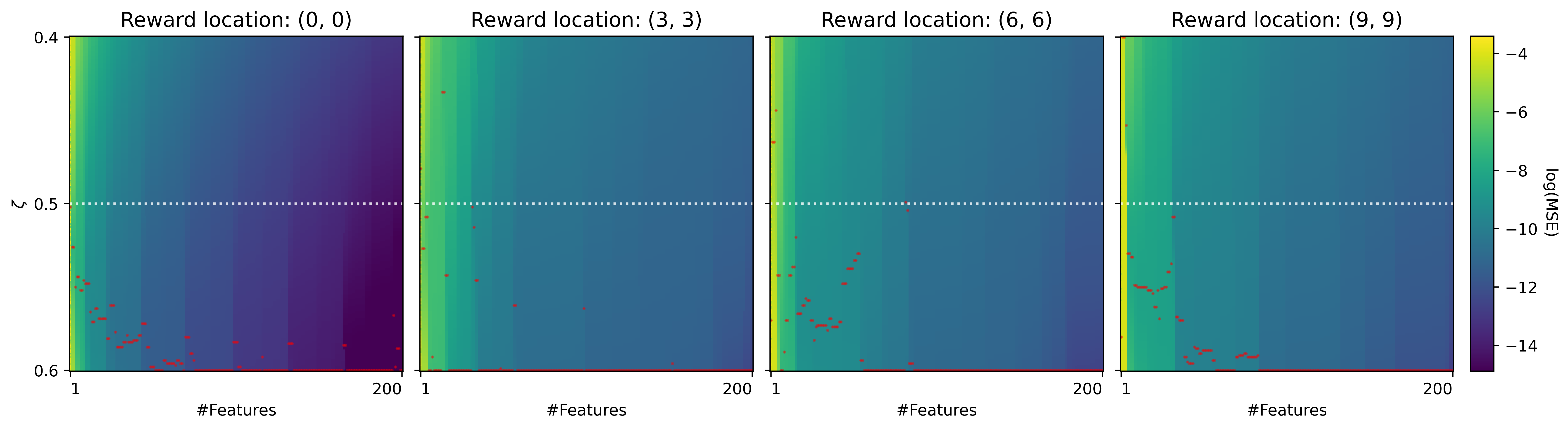}
        \caption{}
    \end{subfigure}
    \vspace{0.4cm}
    \begin{subfigure}{0.77\textwidth}
        \includegraphics[width=1\linewidth]{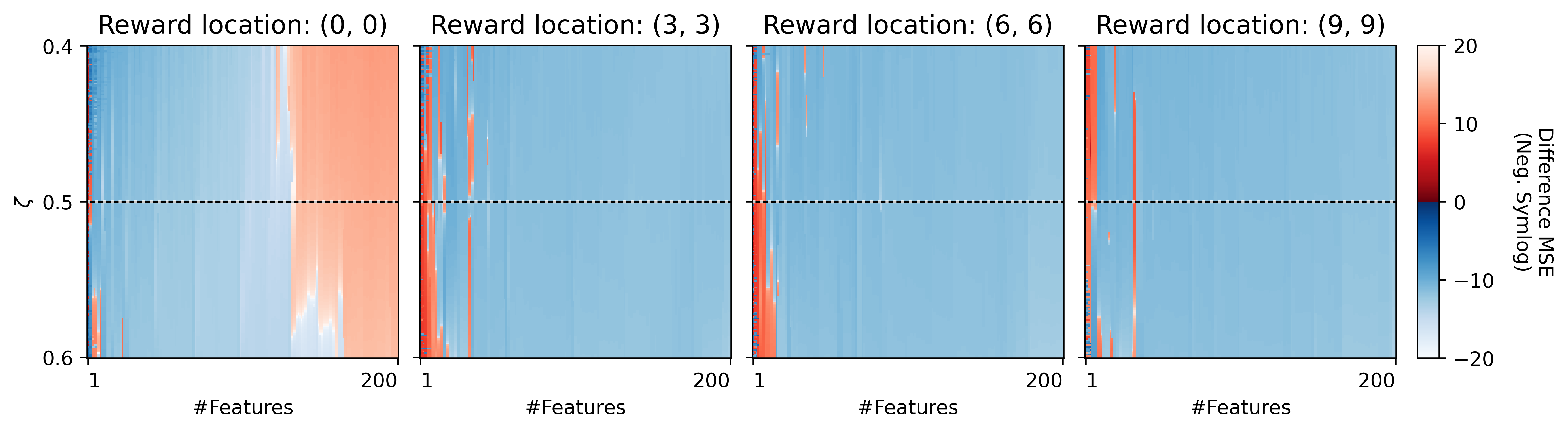}
        \caption{}
    \end{subfigure}
    \caption{Regression results in the 2D environment for $\zeta$-greedy behavior after applying LRA correction (a) with difference visualized in (b).}
    \label{fig:mcsfa_experiments_zeta_2d_constraintrepair}
\end{figure}

Again, the correction mechanism does not improve approximation performance (Figure \ref{fig:mcsfa_experiments_zeta_2d_constraintrepair}), except for the case when a high number of features is used and the reward is located in one corner. Furthermore, goal-averse behavior remains the best choice for all settings.

\begin{figure}[h]
    \centering
    \includegraphics[width=0.9\linewidth]{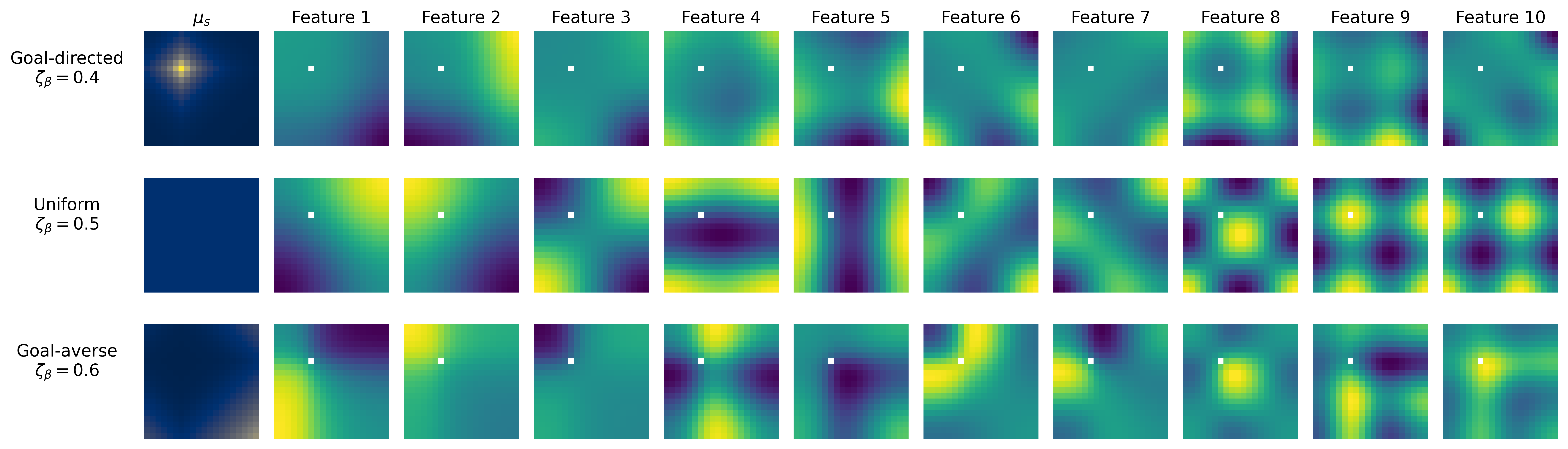}
    \caption{Stationary distribution and example features of Boltzmann behavior for different degrees of goal-directedness or aversion in a 2D lattice environment.}
    \label{fig:mcsfa_features_boltzmann_2d}
\end{figure}

\newpage
\paragraph{Boltzmann behavior} The features resulting from Boltzmann behavior are depicted in Figure \ref{fig:mcsfa_features_boltzmann_2d}. As would be expected, they exhibit a milder scaling effect when compared to $\zeta$-greedy behavior.

In the approximation performance, there is a strongly positive effect visible only when the reward is placed in one corner -- realizing the maximal possible graph distance for the opposite corner and thus the most detrimental scaling. In this case, the difference in behavior to $\zeta$-greedy becomes the largest as the opposite corner realizes the maximum distance to the reward location. Once again, goal-averse behavior performs best across all feature dimensionalities.

\begin{figure}[ht]
    \centering
    \begin{subfigure}{0.77\textwidth}
        \includegraphics[width=1\linewidth]{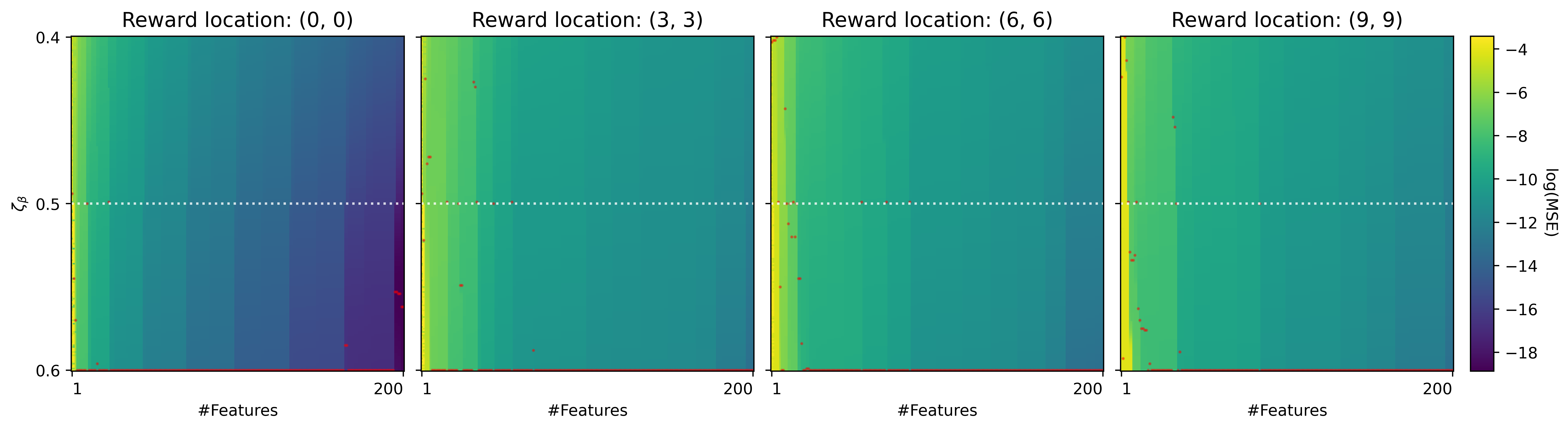} 
        \caption{}
    \end{subfigure}
    \vspace{0.6cm}
    \begin{subfigure}{0.77\textwidth}
        \includegraphics[width=1\linewidth]{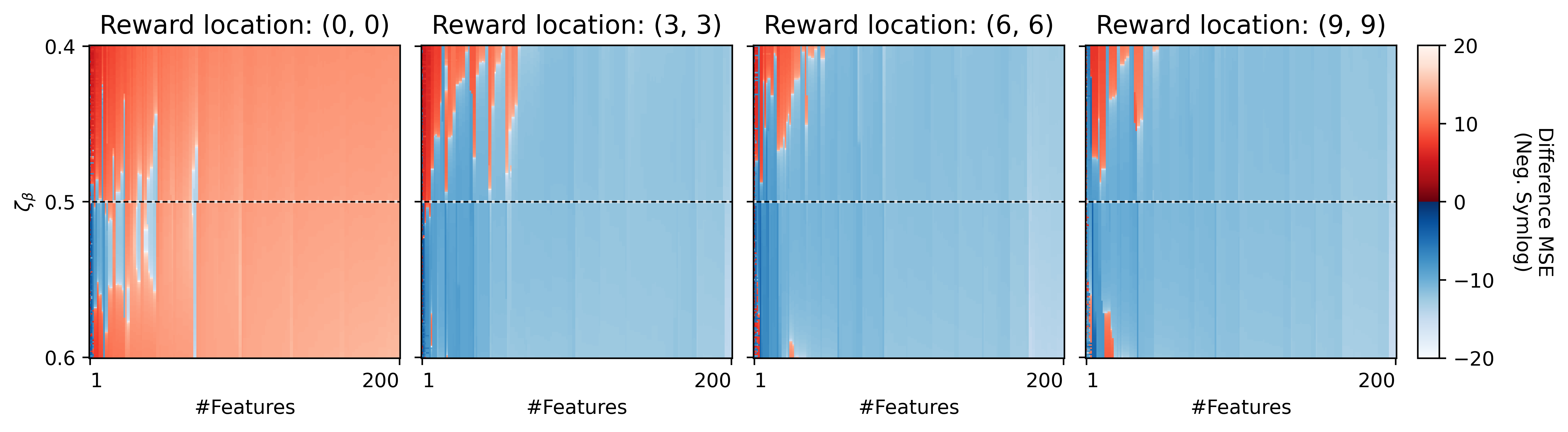} 
        \caption{}
    \end{subfigure}
    \caption{Regression results in the 2D environment for Boltzmann behavior (a) compared with $\zeta$-greedy behavior in (b).}
    \label{fig:mcsfa_experiments_2d_boltzmann}
\end{figure}

\begin{figure}[hp]
    \centering
    \begin{subfigure}{0.9\textwidth}
        \includegraphics[width=1\linewidth]{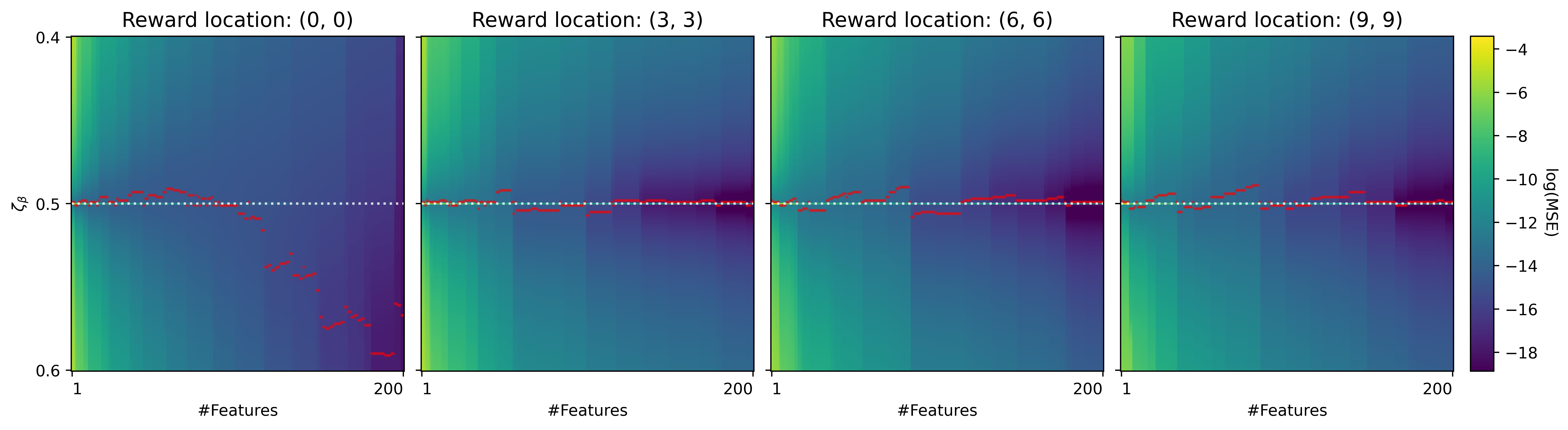} 
        \caption{}
    \end{subfigure}
    \vspace{0.6cm}
    \begin{subfigure}{0.9\textwidth}
        \includegraphics[width=1\linewidth]{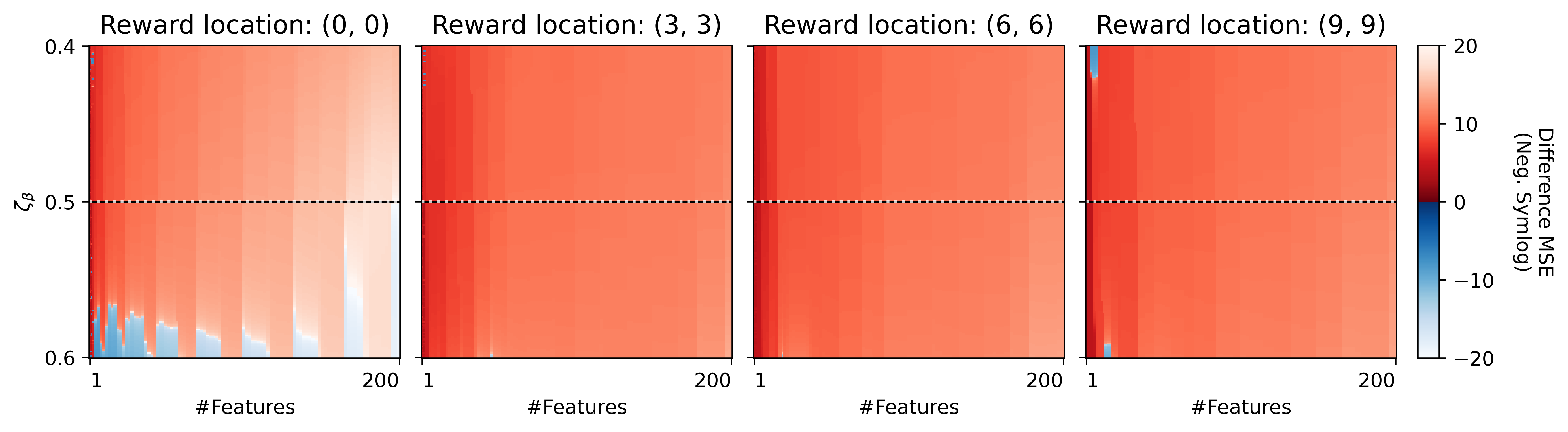} 
        \caption{}
    \end{subfigure}
    \vspace{1cm}
    \begin{subfigure}{0.9\textwidth}
        \includegraphics[width=1\linewidth]{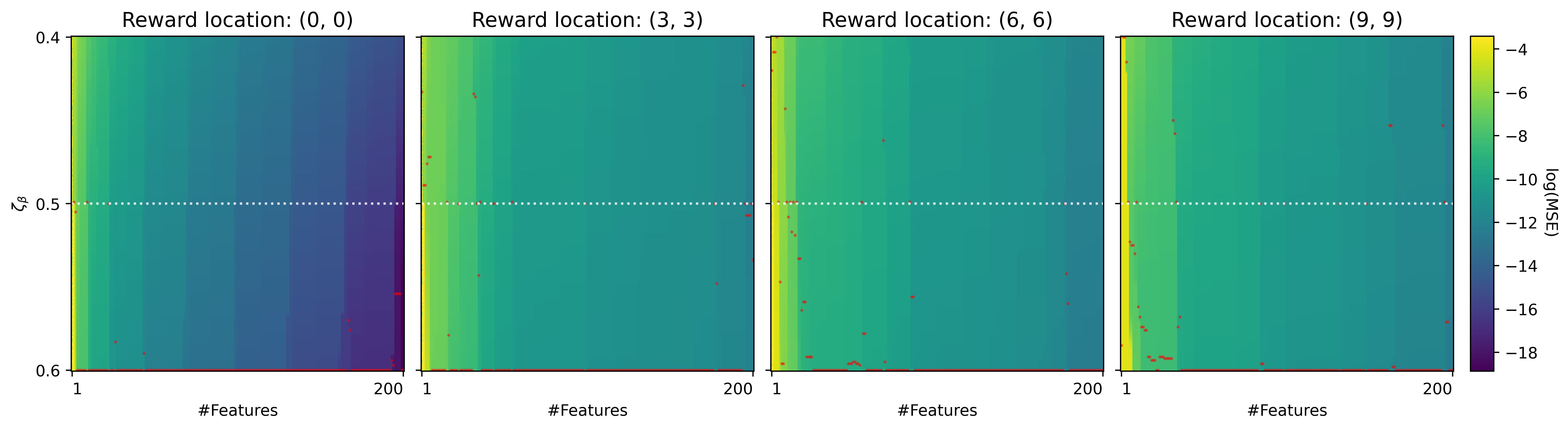} 
        \caption{}
    \end{subfigure}
    \vspace{0.6cm}
    \begin{subfigure}{0.9\textwidth}
        \includegraphics[width=1\linewidth]{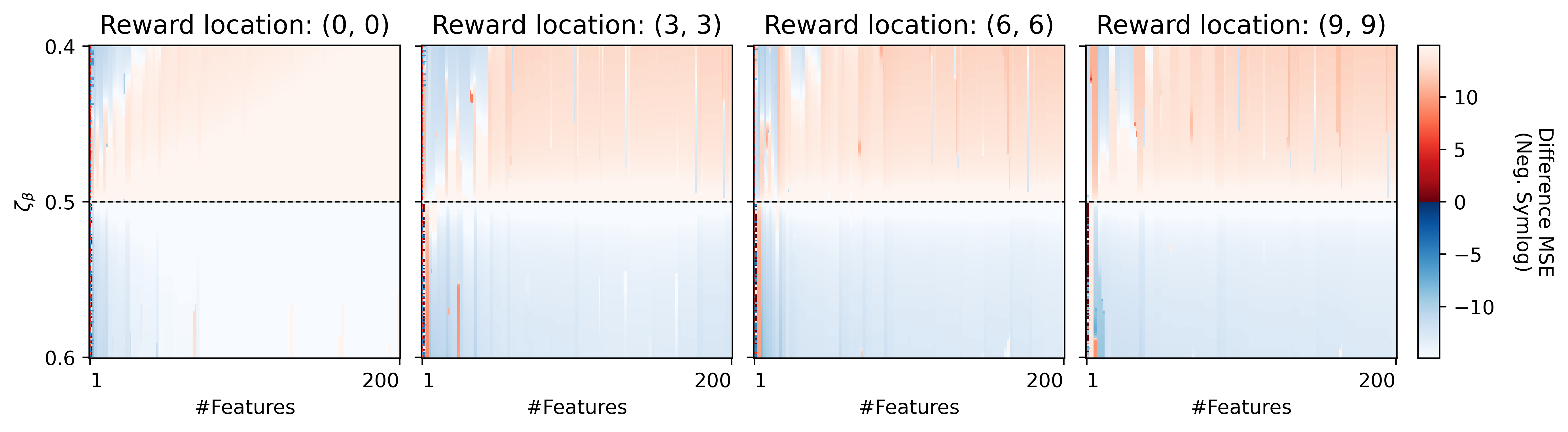} 
        \caption{}
    \end{subfigure}
    \caption{Regression results in the 2D environment for Boltzmann behavior after applying feature scale correction (a) with difference visualized in (b) or after applying LRA correction (c) with difference visualized in (d).}
    \label{fig:mcsfa_experiments_2d_boltzmann_both_repairs}
\end{figure}

Applying the scale correction to the Boltzmann behavior, as seen in Figure \ref{fig:mcsfa_experiments_2d_boltzmann_both_repairs}, turns out to have a strongly positive effect. Scale correction improves performance across essentially all settings of feature dimension, goal-directedness, and reward location. This leads to the best performances stemming from slightly goal-directed or slightly goal-averse behavior.
In contrast, LRA correction only exhibits very small effects, with goal-directed features improving slightly and goal-averse features worsening slightly.

\begin{figure}[h]
    \centering
    \includegraphics[width=0.95\linewidth]{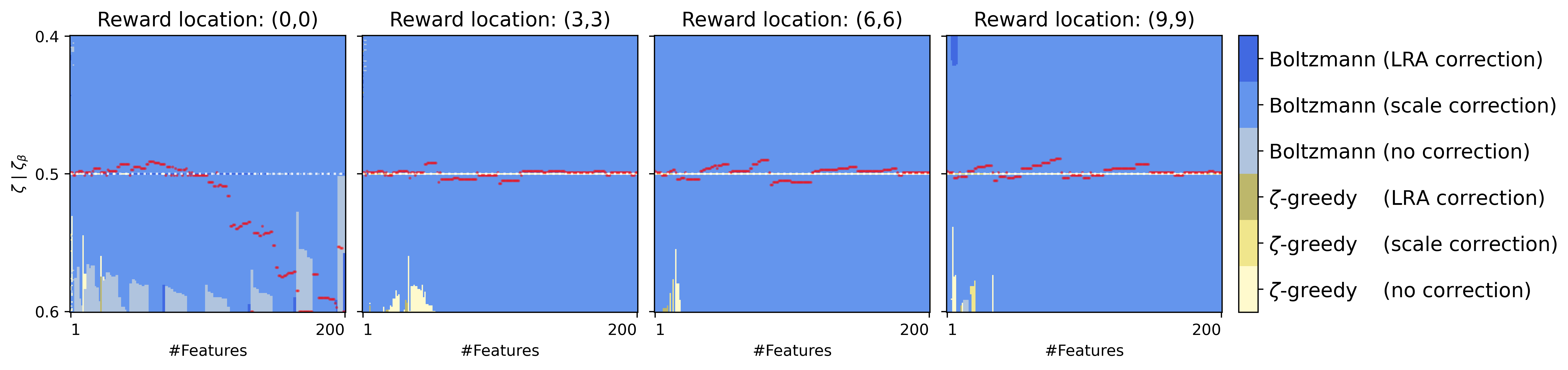}
    \caption{Best performances in the 2D environment between Boltzmann and $\zeta$-greedy behaviors with or without corrections applied.}
    \label{fig:mcsfa_compare_zeta_2d_boltzmann_2d}
\end{figure}

\clearpage
\paragraph{Summary for the 2D lattice environment} Analogously to the comparison for the linear graph environment, all variants of behavior and corrections can be compared, see Figure \ref{fig:mcsfa_compare_zeta_2d_boltzmann_2d}. Regardless of goal-directedness, Boltzmann behavior with scale correction results in the best performance in the overwhelming majority of configurations. This mirrors the results for the linear graph environment, although in that case, the dominance was not as clear.

As opposed to the other settings, this also reduces the relative positive effect of goal-aversion, leading to the best behavior for approximation being close to uniform behavior.

\section{Discussion}
\label{sec:discussion}
This work looks at the effect of using directed behavior to extract optimal slow features. For this, an ergodic Markov chain perspective of slow feature analysis is formulated, and optimal features for this simplified setting are derived in Section \ref{sec:mcsfa}, which confirms a known connection to Laplacian eigenmaps and proto-value functions. 

Optimal features are found to show a strong scaling effect in a spatial environment model, namely directed 1D and 2D lattice graphs, when directedness of the behavior was introduced through a probabilistic $\zeta$-greedy policy that in each state chooses from the set of optimal actions with probability $1-\zeta$ and from the set of non-optimal actions with probability $\zeta$. Optimality was defined in terms of moving toward a reward location in the environment through a reinforcement learning setting.

Goal-directed, goal-averse, and uniform behavior leads to high occupancy around the reward location, low occupancy around the reward location, or uniform occupancy, respectively. Furthermore, optimal features exhibit significantly flattened features in the area of highest occupancy. This confirms previous findings on the influence of the stationary distribution \parencite{Boehmer2013} and scaling effects in continuous settings \parencite{Franzius2007}. Three correction routes are proposed: a behavior modification in the form of Boltzmann behavior, a reformulation of the optimization problem corresponding to learning rate adaptation (LRA) \parencite{Franzius2007}, and a state-wise scale correction of the features according to the occupancy of the state under the stationary distribution.

The evaluation regarding approximation performance of the optimal value function $V^*$ allows the following conclusions for the settings discussed:
\begin{itemize}
    \item Without scale correction or LRA correction, goal-directed behavior leads to features that perform worse in value function approximation when compared to uniform features.
    \item Without scale correction or LRA correction, goal-averse behavior leads to features that perform better in value function approximation when compared to uniform features.
    \item Boltzmann behavior with scale correction leads to better features for approximation in almost all cases, except for the strongest tested goal-aversion in the 1D case. 
    \item LRA correction, as used in this work, in no case leads to the best performance.
\end{itemize}
These results should be viewed in the context of the idealizations and assumptions made. In particular, the following caveats should be considered in their interpretation:
\begin{itemize}
    \item Although a reasonable model, this work simplifies spatially connected environments using finite state spaces and lattice / linear environments.
    \item SFA is typically bound to a fixed family of architectures, and thus can generally not realize optimal features. It is unclear to what extent the features in such a setting exhibit the same effects.
    \item For both, the LRA correction and scale correction, the degree of correction to be applied is a hyperparameter of which only the most natural setting is considered in this work. For example, instead of correcting goal-directed features to be more uniform, an over-correction toward generally better performing goal-averse features is possible.
    \item The ease with which one correction can be applied over the other is not considered in the evaluation. Scale correction requires at least an estimate of the stationary distribution, while LRA correction requires only knowledge of the behavior policy. Furthermore, Boltzmann behavior is widely accepted as a good exploration strategy but is likely to influence reinforcement learning performance through other routes than just approximation performance.
\end{itemize}

Addressing these limitations is not part of this work, but we consider them interesting directions for future research.

\printbibliography

\begin{appendices}
\input{appendix.tex}
\end{appendices}
\end{document}

%% file: gdprocess.tex
\begin{figure}[h]
    \centering
    \begin{tikzpicture}[scale=0.7, every node/.style={scale=0.7}, ->, node distance=2.5cm, state/.style={circle, draw, minimum size=1.2cm}]
  \node[state] (A)                    {$s_0$};
  \node[state]         (B) [right of=A] {$s_1$};
  \node[state, draw=blue!0] (dots) [right of=B] {$\dots$};
  \node[state]         (C) [right of=dots] {$s_{N-2}$};
  \node[state]         (D) [right of=C] {$s_{N-1}$};

  \draw[auto] 
        (A) edge [loop left]node {$1 - \theta$} (A)
        (A) edge [bend left] node {$\theta$} (B) 
        (B) edge [bend left] node {$1 - \theta$} (A)
        
        (B) edge [bend left] node {$\theta$} (dots)
        (dots) edge [bend left] node {$1 - \theta$} (B)
        
        (dots) edge [bend left] node {$\theta$} (C)
        (C) edge [bend left] node {$1 - \theta$} (dots)
        
        (C) edge [bend left] node {$\theta$} (D)
        (D) edge [bend left] node {$1 - \theta$} (C)
        
        (D) edge [loop right] node {$\theta$} (D);
\end{tikzpicture}
\caption{The schematic of a simplified and finite birth-death-process parameterized by a scalar $\theta$.}
    \label{fig:gd_graph}
\end{figure}

%% file: room_graph.tex
\setlength{\belowcaptionskip}{-12pt}
\begin{wrapfigure}{r}{0.4\textwidth}
\centering
 \begin{tikzpicture}[
            > = stealth, 
            auto,
            node distance = 0.7cm, 
            semithick 
        ]

        \tikzstyle{every state}=[
            draw = black,
            thick,
            fill = white,
            minimum size = 4mm,
            scale=0.35
        ]
\node[state] (s0000) [] {};
\node[state] (s0100) [right of=s0000] {};
\node[state] (s0200) [right of=s0100] {};
\node[state] (s0300) [right of=s0200] {};
\node[state] (s0400) [right of=s0300] {};
\node[state] (s0500) [right of=s0400] {};
\node[state] (s0600) [right of=s0500] {};
\node[state] (s0700) [right of=s0600] {};
\node[state] (s0800) [right of=s0700] {};
\node[state] (s0900) [right of=s0800] {};
\node[state] (s0001) [below of=s0000] {};
\node[state] (s0101) [right of=s0001] {};
\node[state] (s0201) [right of=s0101] {};
\node[state] (s0301) [right of=s0201] {};
\node[state] (s0401) [right of=s0301] {};
\node[state] (s0501) [right of=s0401] {};
\node[state] (s0601) [right of=s0501] {};
\node[state] (s0701) [right of=s0601] {};
\node[state] (s0801) [right of=s0701] {};
\node[state] (s0901) [right of=s0801] {};
\node[state] (s0002) [below of=s0001] {};
\node[state] (s0102) [right of=s0002] {};
\node[state] (s0202) [right of=s0102] {};
\node[state] (s0302) [right of=s0202] {};
\node[state] (s0402) [right of=s0302] {};
\node[state] (s0502) [right of=s0402] {};
\node[state] (s0602) [right of=s0502] {};
\node[state] (s0702) [right of=s0602] {};
\node[state] (s0802) [right of=s0702] {};
\node[state] (s0902) [right of=s0802] {};
\node[state] (s0003) [below of=s0002] {};
\node[state] (s0103) [right of=s0003] {};
\node[state] (s0203) [right of=s0103] {};
\node[state] (s0303) [right of=s0203] {};
\node[state] (s0403) [right of=s0303] {};
\node[state] (s0503) [right of=s0403] {};
\node[state] (s0603) [right of=s0503] {};
\node[state] (s0703) [right of=s0603] {};
\node[state] (s0803) [right of=s0703] {};
\node[state] (s0903) [right of=s0803] {};
\node[state] (s0004) [below of=s0003] {};
\node[state] (s0104) [right of=s0004] {};
\node[state] (s0204) [right of=s0104] {};
\node[state] (s0304) [right of=s0204] {};
\node[state] (s0404) [right of=s0304] {};
\node[state] (s0504) [right of=s0404] {};
\node[state] (s0604) [right of=s0504] {};
\node[state] (s0704) [right of=s0604] {};
\node[state] (s0804) [right of=s0704] {};
\node[state] (s0904) [right of=s0804] {};
\node[state] (s0005) [below of=s0004] {};
\node[state] (s0105) [right of=s0005] {};
\node[state] (s0205) [right of=s0105] {};
\node[state] (s0305) [right of=s0205] {};
\node[state] (s0405) [right of=s0305] {};
\node[state] (s0505) [right of=s0405] {};
\node[state] (s0605) [right of=s0505] {};
\node[state] (s0705) [right of=s0605] {};
\node[state] (s0805) [right of=s0705] {};
\node[state] (s0905) [right of=s0805] {};
\node[state] (s0006) [below of=s0005] {};
\node[state] (s0106) [right of=s0006] {};
\node[state] (s0206) [right of=s0106] {};
\node[state] (s0306) [right of=s0206] {};
\node[state] (s0406) [right of=s0306] {};
\node[state] (s0506) [right of=s0406] {};
\node[state] (s0606) [right of=s0506] {};
\node[state] (s0706) [right of=s0606] {};
\node[state] (s0806) [right of=s0706] {};
\node[state] (s0906) [right of=s0806] {};
\node[state] (s0007) [below of=s0006] {};
\node[state] (s0107) [right of=s0007] {};
\node[state] (s0207) [right of=s0107] {};
\node[state] (s0307) [right of=s0207] {};
\node[state] (s0407) [right of=s0307] {};
\node[state] (s0507) [right of=s0407] {};
\node[state] (s0607) [right of=s0507] {};
\node[state] (s0707) [right of=s0607] {};
\node[state] (s0807) [right of=s0707] {};
\node[state] (s0907) [right of=s0807] {};
\node[state] (s0008) [below of=s0007] {};
\node[state] (s0108) [right of=s0008] {};
\node[state] (s0208) [right of=s0108] {};
\node[state] (s0308) [right of=s0208] {};
\node[state] (s0408) [right of=s0308] {};
\node[state] (s0508) [right of=s0408] {};
\node[state] (s0608) [right of=s0508] {};
\node[state] (s0708) [right of=s0608] {};
\node[state] (s0808) [right of=s0708] {};
\node[state] (s0908) [right of=s0808] {};
\node[state] (s0009) [below of=s0008] {};
\node[state] (s0109) [right of=s0009] {};
\node[state] (s0209) [right of=s0109] {};
\node[state] (s0309) [right of=s0209] {};
\node[state] (s0409) [right of=s0309] {};
\node[state] (s0509) [right of=s0409] {};
\node[state] (s0609) [right of=s0509] {};
\node[state] (s0709) [right of=s0609] {};
\node[state] (s0809) [right of=s0709] {};
\node[state] (s0909) [right of=s0809] {};

\path[-] (s0000) edge node {} (s0100);
\path[-] (s0100) edge node {} (s0200);
\path[-] (s0200) edge node {} (s0300);
\path[-] (s0300) edge node {} (s0400);
\path[-] (s0400) edge node {} (s0500);
\path[-] (s0500) edge node {} (s0600);
\path[-] (s0600) edge node {} (s0700);
\path[-] (s0700) edge node {} (s0800);
\path[-] (s0800) edge node {} (s0900);
\path[-] (s0000) edge node {} (s0001);
\path[-] (s0001) edge node {} (s0101);
\path[-] (s0100) edge node {} (s0101);
\path[-] (s0101) edge node {} (s0201);
\path[-] (s0200) edge node {} (s0201);
\path[-] (s0201) edge node {} (s0301);
\path[-] (s0300) edge node {} (s0301);
\path[-] (s0301) edge node {} (s0401);
\path[-] (s0400) edge node {} (s0401);
\path[-] (s0401) edge node {} (s0501);
\path[-] (s0500) edge node {} (s0501);
\path[-] (s0501) edge node {} (s0601);
\path[-] (s0600) edge node {} (s0601);
\path[-] (s0601) edge node {} (s0701);
\path[-] (s0700) edge node {} (s0701);
\path[-] (s0701) edge node {} (s0801);
\path[-] (s0800) edge node {} (s0801);
\path[-] (s0801) edge node {} (s0901);
\path[-] (s0900) edge node {} (s0901);
\path[-] (s0001) edge node {} (s0002);
\path[-] (s0002) edge node {} (s0102);
\path[-] (s0101) edge node {} (s0102);
\path[-] (s0102) edge node {} (s0202);
\path[-] (s0201) edge node {} (s0202);
\path[-] (s0202) edge node {} (s0302);
\path[-] (s0301) edge node {} (s0302);
\path[-] (s0302) edge node {} (s0402);
\path[-] (s0401) edge node {} (s0402);
\path[-] (s0402) edge node {} (s0502);
\path[-] (s0501) edge node {} (s0502);
\path[-] (s0502) edge node {} (s0602);
\path[-] (s0601) edge node {} (s0602);
\path[-] (s0602) edge node {} (s0702);
\path[-] (s0701) edge node {} (s0702);
\path[-] (s0702) edge node {} (s0802);
\path[-] (s0801) edge node {} (s0802);
\path[-] (s0802) edge node {} (s0902);
\path[-] (s0901) edge node {} (s0902);
\path[-] (s0002) edge node {} (s0003);
\path[-] (s0003) edge node {} (s0103);
\path[-] (s0102) edge node {} (s0103);
\path[-] (s0103) edge node {} (s0203);
\path[-] (s0202) edge node {} (s0203);
\path[-] (s0203) edge node {} (s0303);
\path[-] (s0302) edge node {} (s0303);
\path[-] (s0303) edge node {} (s0403);
\path[-] (s0402) edge node {} (s0403);
\path[-] (s0403) edge node {} (s0503);
\path[-] (s0502) edge node {} (s0503);
\path[-] (s0503) edge node {} (s0603);
\path[-] (s0602) edge node {} (s0603);
\path[-] (s0603) edge node {} (s0703);
\path[-] (s0702) edge node {} (s0703);
\path[-] (s0703) edge node {} (s0803);
\path[-] (s0802) edge node {} (s0803);
\path[-] (s0803) edge node {} (s0903);
\path[-] (s0902) edge node {} (s0903);
\path[-] (s0003) edge node {} (s0004);
\path[-] (s0004) edge node {} (s0104);
\path[-] (s0103) edge node {} (s0104);
\path[-] (s0104) edge node {} (s0204);
\path[-] (s0203) edge node {} (s0204);
\path[-] (s0204) edge node {} (s0304);
\path[-] (s0303) edge node {} (s0304);
\path[-] (s0304) edge node {} (s0404);
\path[-] (s0403) edge node {} (s0404);
\path[-] (s0404) edge node {} (s0504);
\path[-] (s0503) edge node {} (s0504);
\path[-] (s0504) edge node {} (s0604);
\path[-] (s0603) edge node {} (s0604);
\path[-] (s0604) edge node {} (s0704);
\path[-] (s0703) edge node {} (s0704);
\path[-] (s0704) edge node {} (s0804);
\path[-] (s0803) edge node {} (s0804);
\path[-] (s0804) edge node {} (s0904);
\path[-] (s0903) edge node {} (s0904);
\path[-] (s0004) edge node {} (s0005);
\path[-] (s0005) edge node {} (s0105);
\path[-] (s0104) edge node {} (s0105);
\path[-] (s0105) edge node {} (s0205);
\path[-] (s0204) edge node {} (s0205);
\path[-] (s0205) edge node {} (s0305);
\path[-] (s0304) edge node {} (s0305);
\path[-] (s0305) edge node {} (s0405);
\path[-] (s0404) edge node {} (s0405);
\path[-] (s0405) edge node {} (s0505);
\path[-] (s0504) edge node {} (s0505);
\path[-] (s0505) edge node {} (s0605);
\path[-] (s0604) edge node {} (s0605);
\path[-] (s0605) edge node {} (s0705);
\path[-] (s0704) edge node {} (s0705);
\path[-] (s0705) edge node {} (s0805);
\path[-] (s0804) edge node {} (s0805);
\path[-] (s0805) edge node {} (s0905);
\path[-] (s0904) edge node {} (s0905);
\path[-] (s0005) edge node {} (s0006);
\path[-] (s0006) edge node {} (s0106);
\path[-] (s0105) edge node {} (s0106);
\path[-] (s0106) edge node {} (s0206);
\path[-] (s0205) edge node {} (s0206);
\path[-] (s0206) edge node {} (s0306);
\path[-] (s0305) edge node {} (s0306);
\path[-] (s0306) edge node {} (s0406);
\path[-] (s0405) edge node {} (s0406);
\path[-] (s0406) edge node {} (s0506);
\path[-] (s0505) edge node {} (s0506);
\path[-] (s0506) edge node {} (s0606);
\path[-] (s0605) edge node {} (s0606);
\path[-] (s0606) edge node {} (s0706);
\path[-] (s0705) edge node {} (s0706);
\path[-] (s0706) edge node {} (s0806);
\path[-] (s0805) edge node {} (s0806);
\path[-] (s0806) edge node {} (s0906);
\path[-] (s0905) edge node {} (s0906);
\path[-] (s0006) edge node {} (s0007);
\path[-] (s0007) edge node {} (s0107);
\path[-] (s0106) edge node {} (s0107);
\path[-] (s0107) edge node {} (s0207);
\path[-] (s0206) edge node {} (s0207);
\path[-] (s0207) edge node {} (s0307);
\path[-] (s0306) edge node {} (s0307);
\path[-] (s0307) edge node {} (s0407);
\path[-] (s0406) edge node {} (s0407);
\path[-] (s0407) edge node {} (s0507);
\path[-] (s0506) edge node {} (s0507);
\path[-] (s0507) edge node {} (s0607);
\path[-] (s0606) edge node {} (s0607);
\path[-] (s0607) edge node {} (s0707);
\path[-] (s0706) edge node {} (s0707);
\path[-] (s0707) edge node {} (s0807);
\path[-] (s0806) edge node {} (s0807);
\path[-] (s0807) edge node {} (s0907);
\path[-] (s0906) edge node {} (s0907);
\path[-] (s0007) edge node {} (s0008);
\path[-] (s0008) edge node {} (s0108);
\path[-] (s0107) edge node {} (s0108);
\path[-] (s0108) edge node {} (s0208);
\path[-] (s0207) edge node {} (s0208);
\path[-] (s0208) edge node {} (s0308);
\path[-] (s0307) edge node {} (s0308);
\path[-] (s0308) edge node {} (s0408);
\path[-] (s0407) edge node {} (s0408);
\path[-] (s0408) edge node {} (s0508);
\path[-] (s0507) edge node {} (s0508);
\path[-] (s0508) edge node {} (s0608);
\path[-] (s0607) edge node {} (s0608);
\path[-] (s0608) edge node {} (s0708);
\path[-] (s0707) edge node {} (s0708);
\path[-] (s0708) edge node {} (s0808);
\path[-] (s0807) edge node {} (s0808);
\path[-] (s0808) edge node {} (s0908);
\path[-] (s0907) edge node {} (s0908);
\path[-] (s0008) edge node {} (s0009);
\path[-] (s0009) edge node {} (s0109);
\path[-] (s0108) edge node {} (s0109);
\path[-] (s0109) edge node {} (s0209);
\path[-] (s0208) edge node {} (s0209);
\path[-] (s0209) edge node {} (s0309);
\path[-] (s0308) edge node {} (s0309);
\path[-] (s0309) edge node {} (s0409);
\path[-] (s0408) edge node {} (s0409);
\path[-] (s0409) edge node {} (s0509);
\path[-] (s0508) edge node {} (s0509);
\path[-] (s0509) edge node {} (s0609);
\path[-] (s0608) edge node {} (s0609);
\path[-] (s0609) edge node {} (s0709);
\path[-] (s0708) edge node {} (s0709);
\path[-] (s0709) edge node {} (s0809);
\path[-] (s0808) edge node {} (s0809);
\path[-] (s0809) edge node {} (s0909);
\path[-] (s0908) edge node {} (s0909);
    \end{tikzpicture}
    \caption{Example of a lattice with $10\times10$ states}
    \label{fig:single_room_graph}
\end{wrapfigure}
\setlength{\belowcaptionskip}{0pt}

%% file: appendix.tex
\section{Matrix Derivatives for SFA on Markov Chains}
\label{appendix:matrix_derivatives}
Even basic calculus involving matrices can sometimes pose to be elusive and hard to parse. This is why we will include some useful derivations here in extensive detail, which have been used in Section \ref{sec:mcsfa}.

We use the convention to write the derivative of a scalar $y$ with respect to a matrix $\mathbf{X}=(X_{ij})_{ij}$ is again a matrix 

\begin{equation}
    \frac{\partial y}{\partial \mathbf{X}} = \Big( 
    \frac{\partial y}{\partial X_{ij}} \Big)_{ij}
\end{equation}

of similar dimensions and entries corresponding to partial derivatives of the entries of $\mathbf{X}$.

Some useful identities:
\begin{equation}
\frac{\partial \Tr(\mathbf{A}^T\mathbf{BA})}{\partial \mathbf{A}} = \Big(\frac{\partial \Tr(\mathbf{A}^T\mathbf{BA})}{\partial A_{ij}}\Big)_{ij} 
\end{equation}
and for individual entries
\begin{align}
\frac{\partial \Tr(\mathbf{A}^T\mathbf{BA})}{\partial A_{ij}} &= \frac{\partial }{\partial A_{ij}} \Tr(\mathbf{A}^T\mathbf{BA}) \\
&= \frac{\partial }{\partial A_{ij}} \sum_u \sum_l \sum_n B_{ln}A_{lu}A_{nu}
\\
&= \underbrace{\frac{\partial }{\partial A_{ij}} \Big(\sum_{u\neq j}\sum_l \sum_n B_{ln}A_{lu}A_{nu}\Big)}_{=0} 
+ \frac{\partial }{\partial A_{ij}} \Big(\sum_l \sum_n B_{ln}A_{lj}A_{nj}\Big) \\
&= \frac{\partial }{\partial A_{ij}} \Big(\sum_l \sum_n B_{ln}A_{lj}A_{nj}\Big) \\
&= \frac{\partial }{\partial A_{ij}} \Big(\underbrace{\sum_n B_{in}A_{ij}A_{nj}}_{l=i} + \underbrace{\sum_{l\neq i} \sum_n B_{ln}A_{lj}A_{nj}}_{l\neq i}\Big)
\\
&= \frac{\partial }{\partial A_{ij}} \Big(\underbrace{\vphantom{\sum_{n \neq i}}B_{ii}A_{ij}A_{ij}}_{l=i,n=i}+\underbrace{\sum_{n\neq i} B_{in}A_{ij}A_{nj}}_{l=i, n\neq i} + 
\underbrace{\sum_{l\neq i} B_{li}A_{lj}A_{ij}}_{l\neq i, n = i}
+ \underbrace{\sum_{l\neq i} \sum_{n\neq i} B_{ln}A_{lj}A_{nj}}_{l\neq i, n\neq i}\Big)\\
&= 2B_{ii}A_{ij} + \sum_{n\neq i} B_{in}A_{nj} + \sum_{l\neq i} B_{li}A_{lj} + 0\\
&= \sum_{n} B_{in}A_{nj} + \sum_{l} B_{li}A_{lj}
=\mathbf{A}_{\cdot j} \mathbf{B}_{i\cdot} + \mathbf{A}_{\cdot j} \mathbf{B}_{\cdot i} \label{eq:basic_trace_derivative_elementwise}
\end{align}
where $\mathbf{B}_{i\cdot}$ and $\mathbf{B}_{\cdot i}$ are the row $i$ or column $i$ of the matrix $\mathbf{B}$ as vector, respectively, and similar for $\mathbf{A}_{\cdot j}$ and the products in the last term are inner products.
Thus, the full matrix derivative can be written as
\begin{align}
    \frac{\partial \Tr(\mathbf{A}^T\mathbf{BA})}{\partial \mathbf{A}} = \mathbf{B}\mathbf{A} + \mathbf{B}^T\mathbf{A}
\end{align}
or, in the case that $\mathbf{B}$ is symmetric, as
\begin{align}
    \frac{\partial \Tr(\mathbf{A}^T\mathbf{BA})}{\partial \mathbf{A}} = 2\mathbf{B}\mathbf{A}. \label{eq:trace_derivative}
\end{align}

\noindent Another identity used, for a diagonal matrix $\mathbf{\Lambda}$ with diagonal entries $\lambda_i$, is
\begin{align}
    \frac{\partial \Tr(\mathbf{\Lambda}(\mathbf{A}^T\mathbf{BA} - \mathbf{I}))}{\partial A_{ij}} &=
    \frac{\partial \Tr(\mathbf{\Lambda}\mathbf{A}^T\mathbf{BA} - \mathbf{\Lambda})}{\partial A_{ij}}\\
    &=
    \frac{\partial}{\partial A_{ij}}\Tr(\mathbf{\Lambda}\mathbf{A}^T\mathbf{BA}) - 
    \underbrace{\frac{\partial}{\partial A_{ij}}\Tr(\mathbf{\Lambda})}_{=0}\\
    &= \frac{\partial }{\partial A_{ij}} \sum_u \sum_l \sum_n \lambda_u B_{ln}A_{lu}A_{nu}\\
    &= \underbrace{\frac{\partial }{\partial A_{ij}} \Big(\sum_{u\neq j}\sum_l \sum_n \lambda_u B_{ln}A_{lu}A_{nu}\Big)}_{=0} 
+ \frac{\partial }{\partial A_{ij}} \Big(\sum_l \sum_n \lambda_j B_{ln}A_{lj}A_{nj}\Big) \\
    &= \frac{\partial }{\partial A_{ij}} \sum_l \sum_n \lambda_j B_{ln}A_{lj}A_{nj} \\
    &=\lambda_j \frac{\partial }{\partial A_{ij}} \sum_l \sum_n  B_{ln}A_{lj}A_{nj} \\
    &\stackrel{\ref{eq:basic_trace_derivative_elementwise}}{=}\lambda_j \mathbf{A}_{\cdot j} \mathbf{B}_{i\cdot} + \lambda_j \mathbf{A}_{\cdot j} \mathbf{B}_{\cdot i} 
\end{align}
The full matrix derivative can thus be written as
\begin{align}
\frac{\partial \Tr(\mathbf{\Lambda}(\mathbf{A}^T\mathbf{BA} - \mathbf{I}))}{\partial \mathbf{A}} = 
\mathbf{BA\Lambda} +\mathbf{B}^T\mathbf{A\Lambda} = 
(\mathbf{B} +\mathbf{B}^T)\mathbf{A\Lambda} 
\end{align}
or, in the case of a symmetric matrix $\mathbf{B}$, as \begin{align}
\frac{\partial \Tr(\mathbf{\Lambda}(\mathbf{A}^T\mathbf{BA} - \mathbf{I}))}{\partial \mathbf{A}} = 
2\mathbf{B}\mathbf{A \Lambda} \label{eq:trace_derivative_diagonal}
\end{align}